\documentclass[3p,times]{elsarticle}
\usepackage{epstopdf}

\journal{Journal of \LaTeX\ Templates}

\usepackage[cmex10]{amsmath}
\usepackage{array}
\usepackage{mdwmath}
\usepackage{mdwtab}
\newcommand{\etal}{\textit{et al}.}

\begin{document}

\begin{frontmatter}

\title{Joint direct estimation of 3D geometry and 3D motion using spatio temporal gradients}


\author[ugraddress]{Francisco~Barranco\corref{mycorrespondingauthor}}
\cortext[mycorrespondingauthor]{Corresponding author} \ead{fbarranco@ugr.es}
\author[umdaddress]{Cornelia~Ferm\"uller}
\author[umdaddress]{Yiannis~Aloimonos}
\author[ugraddress]{Eduardo~Ros}

\address[ugraddress]{Dept. of Computer Architecture and Technology, CITIC, University of Granada, Spain.}
\address[umdaddress]{Dept. of Computer Science, UMIACS, University of Maryland, College Park, MD, USA.}

\tnotetext[mytitlenote]{\textcopyright 2018. This manuscript version is made available under the CC-BY-NC-ND 4.0 license \textit{http://creativecommons.org/licenses/by-nc-nd/4.0/}}

\begin{abstract}
Conventional image motion based structure from motion methods first compute optical flow, then solve for the 3D motion parameters based on the epipolar constraint, and finally recover the 3D geometry of the scene. However, errors in optical flow due to regularization can lead to large errors in 3D motion and structure. This paper investigates whether performance and consistency can be improved by avoiding optical flow estimation in the early stages of the structure from motion pipeline, and it proposes a new direct method based on image gradients (normal flow) only. The main idea lies in a reformulation of the positive-depth constraint, which allows the use of well-known minimization techniques to solve for 3D motion. The 3D motion estimate is then refined and structure estimated adding a regularization based on depth.
Experimental comparisons on standard synthetic datasets and the real-world driving benchmark dataset KITTI using three different optic flow algorithms show that the method achieves better accuracy in all but one case. Furthermore, it outperforms existing normal flow based 3D motion estimation techniques. Finally, the recovered 3D geometry is shown to be also very accurate.
\end{abstract}


\end{frontmatter}


\section{Introduction}
\label{sec:1}
The problem of egomotion or self-motion estimation from a moving monocular observer, after many years of research, is still considered a difficult problem. Recently it has attracted renewed attention in the Computer Vision community due to emerging  applications in robotics, autonomous navigation and augmented reality. Physically, the motion of the camera can be interpreted as the linear combination of a 3D translation followed by a 3D rotation. The instantaneous motion captured contains information about the camera's 3D motion and the 3D scene geometry. Egomotion estimation amounts to computing five parameters: three for the 3D rotation and two for the axis of the 3D translation, because without additional information, there is an ambiguity between translational velocity and depth. Based on the 3D motion, the 3D relative structure can be estimated.

The classic approach to estimating structure and motion employs three steps: first, the full dense optical flow between sucessive frames is estimated; second, the 3D translation and 3D rotation are recovered using the optical flow, possibly making assumptions about the camera motion and the scene; third, the 3D geometry up to the scaling factor is estimated \cite{raudies_efficient_2009, raudies_review_2012, pauwels_optimal_2006,zhang_consistency_2002}. Instead of dense flow, a sparse set of feature correspondences is often used, as is common in standard visual odometry and SLAM methods \cite{davison2007monoslam, mur_orb2_2017, forster_svo_2014}. However, recent SLAM formulations \cite{ engel_direct_2017} do not estimate 3D motion through constraints independent of depth, but estimate 3D motion and depth combined by minimizing photometric/geometric distance that explain image patch matches. The focus of this paper is on the evaluation of depth independent constraints.

The main constraint to estimate 3D motion from video independent of structure, is the epipolar contraint. However, it requires as input optical flow or correspondences. One problem is that optical flow cannot be estimated accurately. Most top optical flow techniques are based on the work of Horn \& Schunk \cite{horn_determining_1981}. The key assumption is that the change of the intensity over a small time interval remains constant. Since this only provides one equation and flow fields are two-dimensional,  additional constraints on the flow field are enforced assuming a smooth variation of the field spatially in local neighborhoods \cite{brox_large_2011, vogel_evaluation_2013, sun_quantitative_2014}. These assumptions cause the optical flow to be imprecise at object contours, where there are occlusions or when the motion is large. Motion fields do not vary smoothly close to object boundaries, occlusions cause mismatches, and large motions violate the assumption of local intensity constancy.

Another motion constraint, independent of structure, is the depth positivity constraint \cite{fermuller_passive_1995,fermuller_observability_2000}, also referred to as cheirality constraint \cite{hartley_multiple_2003}. The scene has to be in front of the camera, and thus the depth has to be positive. This constraint can be used directly from  \textit{"measurable image quantities"}, i.e. the spatial and temporal image intensity gradients, or normal flow, in so-called direct methods.
The positivity constraint is the only constraint applicable to normal flow without making assumptions on scene depth or shape. Although, it can also be applied to optical flow. It has not been popular, because  there has not been a formulation allowing to implement the depth positivity inequality in an efficient way. This work proposes such a formulation of  depth positivity, borrowing from current machine-learning models. On its basis then a new direct method for estimation of 3D motion and structure is proposed.

The main contributions of this paper are: 1) a new formulation of the depth positivity constraint that allows for efficient minimization; 2) a new direct method for egomotion and 3D structure estimation from normal flow only; 3) a comparison of the method  to a number of studies using optical flow and to other direct approaches using normal flow on a number of datasets including real-world sequences, which show that the method outperforms previous methods using different error metrics.

\section{Related methods}
\label{sec:2}
Longuet-Higgins in his seminal work presented the constraints for computing egomotion from image measurements \cite{longuet_computer_1987}. Classic approaches that use optical flow, model the 3D rigid displacement between frames with instanatneous 3D velocity \cite{horn1986robot}. Bruss and Horn \cite{bruss_passive_1983} used the weighted bilinear constraint and solved for the egomotion using a least-squares optimization. 
Kanatani \cite{kanatani_3dinterpretation_1993} proposed a method based on the differential epipolar constraints that accounts and corrects for the bias in the translational velocity.
Modern optimal methods compute more accurate solutions at the expense of nonlinear objective functions and the risk of falling into local minima \cite{pauwels_optimal_2006}. Zhang \cite{zhang_consistency_2002} considered a two-step iterative approach that estimates the translation separately from the rotation, using numerical Gauss-Newton approximation. Pauwels and van Halle \cite{pauwels_optimal_2006} proposed a method that integrates depth cues into the minimization of the squared-distance image reprojection error. The method improved previous works by reducing the risk of falling into local minima. Instead of using random initializations or initializations from classical linear method solutions, the authors used a weighting strategy to change the complexity at each iteration, making the algorithm more robust.
Raudies and Neumann \cite{raudies_efficient_2009,raudies_review_2012} also used a polynomial solution for the bilinear constraint employing auxiliary variables (as done in \cite{kanatani_3dinterpretation_1993}) for a more efficient method of linear complexity; it applies RANSAC to reduce the impact of noise as well. For a complete review on 3D motion estimation methods see \cite{raudies_review_2012}.

On the other hand, many methods are feature-based. During the last years in the context of autonomous navigation, most 3D motion approaches proposed are SLAM methods. The most recent work from Engel \etal \cite{engel_direct_2017} presented a real-time method for visual odometry that jointly estimates egomotion and geometry without using a smoothing prior, but by sampling a set of locations throughout the sequence. The same authors developed LSD-SLAM \cite{engel_lsd_2014}, which uses a photometric error measure and a geometric smoothing prior to solve the optimization on a full dense optical flow field throughout consecutive frames. Mur-Artal \etal \cite{mur_orb2_2017} presented an open-source framework that allows for odometry estimation, using a geometric distance measure without geometric prior. Ranftl \etal \cite{ranftl_dense_2016} proposed a monocular depth estimation approach that jointly solves for egomotion using a full dense regularized optical flow field assuming its smoothness. Finally, Forster \etal \cite{forster_svo_2014} also described a probabilistic hybrid approach for semi-direct visual odometry for UAVs, that uses a formulation based on a geometric prior for the initialization but avoids its use in the subsequent model optimization.


Several works in the literature have studied direct methods that use normal flow avoiding optical flow estimation.
Sinclair \etal \cite{sinclair_robust_1994} proposed a normal flow based method, but only considered translation.
Ferm\"uller and Aloimonos developed methods \cite{fermuller_passive_1995, fermuller_qualitative_1995} that formulate 3D motion estimation as a pattern recognition problem. Subsets of normal flow vectors define global patterns in the image plane whose  location encodes the egomotion parameters. Finally, Brodsky \etal \cite{brodsky1998shape} estimated 3D motion from normal flow assuming planar patches, and Ji and Ferm\"uller based on this method proposed a solution \cite{ji_3dshape_2006} that combines the estimated motion fields from consecutive frames using a constraints on the inverse depth for regular image patches and for segmented regions.


More recently, Hui and Chung \cite{hui_structure_2013} used a binocular sequence but computed monocular motion with normal flows and then estimated depth from the stereo system without explicitly computing correspondences. They optimized for the 3D motion using the so-called Apparent Flow Direction (AFD) and Apparent Flow Magnitude (AFM) constraints. The first constraint, AFD, is a relaxation of the positive depth constraint. The second constraint, AFM, affects the motion magnitude and only uses the normal flows orthogonal to the translational component. The same authors in \cite{hui_determining_2015} presented a more comprehensive evaluation and compared their results with optical flow methods. In \cite{yuan_direct_2013} the authors described a two-step method that first estimates the 3D motion using  only the normal flow vectors orthogonal to the translational motion component. Next the method solves for the   rotation discarding the solutions that lie out of the half-plane consistent with the normal flow estimates, defined by the first step. In \cite{yuan_camera_2015} the same authors utilize k-means clustering to group the flow vectors that support the same FOE candidate. Next, the rotational parameters are estimated using RANSAC and finally, a confirmation strategy based on the half-plane constraint helps finding potential solutions. Finally, using the same normal flow constraints, other works extract the 3D structure from the scene \cite{hui_dense_2014, hui_structure_2013}.

\section{Egomotion estimation from  normal flow}
\label{sec:3}
Let us first rephrase the egomotion problem definition as follows: Recover the 3D trajectory and pose of a monocular observer undergoing a rigid motion in a stationary environment.
Assuming that the coordinate system of the camera is centered in the principal point and $f$ is the focal length, the 3D velocity  $\textbf{v}$ of a point $\textbf{x}$ is defined as $\textbf{v} = -\textbf{t} - \textbf{w} \times \textbf{x}$, where $\textbf{t}=(t_x, t_y, t_z)^{T}$ is the velocity of translation,  $\textbf{w}=(w_x, w_y, w_z)^{T}$ is the velocity of rotation, and $\times$ represents the cross product. Then, the motion field $\textbf{u}(\textbf{x})=(u, v)^T$ at location $\textbf{x}=(x,y)^{T}$ is related to  the 3D velocities \cite{longuet_interpretation_1980} as
\begin{eqnarray}
\label{eq:longuet}
\textbf{u}(\textbf{x}) = \frac{1}{Z(\textbf{x})}~A(\textbf{x})~\textbf{t} + B(\textbf{x})~\textbf{w}
\end{eqnarray}
with
\begin{equation}
\begin{split}
A(\textbf{x}) =
\begin{bmatrix}
    -f & 0 & x \\
    0 & -f & y
\end{bmatrix}
, \\
B(\textbf{x}) =
\begin{bmatrix}
    \frac{x y}{f} & \left(-\frac{x^2}{f}-f\right) & y \\
    \left(\frac{y^2}{f} + f\right) & -\frac{x y}{f} & -x
\end{bmatrix}
\end{split}
\label{eq:longuet_2}
\end{equation}
where the motion is expressed as the sum  of the rotational component $B(\textbf{x})\textbf{w}$, and the translational component $\frac{1}{Z(\textbf{x})}A(\textbf{x})\textbf{t}$, which  also depends on the depth of the scene $Z$. Due to this dependency there are five parameters to be estimated, namely the translational velocity axis $(t_x/t_z,t_y/t_z)^{T}$ and the rotational velocity $(w_x, w_y, w_z)^{T}$. 

\subsection{Normal flow}
\label{subsec:3.1}
To estimate motion fields, we use  the so-called \textit{optical flow constraint} (see \cite{horn_determining_1981}), which assumes that the intensity $I$ at a point remains constant over a short time interval $\delta t$. Approximating the image brightness function with  a first order Taylor expansion, we obtain Eq.~(\ref{eq:OFC})
\begin{equation}
0 = I(x,y,t) - I(x + u \delta t,\ y + v \delta t,\ t+\delta t) \approx I_x u + I_y v + I_t,
\label{eq:OFC}
\end{equation}

where  $I_x, I_y, I_t$ are the partial derivatives of the brightness, and $\textbf{u} = (u, v)$ is the image motion. Equation \ref{eq:OFC} provides one constraint and defines the flow component parallel to the gradient. To obtain a second component, additional assumptions, such as smoothness of the flow or a parametric model in the image coordinates are assumed.

Direct methods do not require computing the two-dimensional optical flow, but instead use directly the image gradients as input (equation \ref{eq:OFC}). For a geometric interpretation, we consider the projection of the optical flow on the gradient direction, called the \textit{normal flow}, which amounts to:
\begin{equation}
	\textbf{u}_\textbf{n}(\textbf{x}) = \frac{-I_t(\textbf{x})}{\left\|\nabla I(\textbf{x})\right\|^{2}} \nabla I(\textbf{x})
	\label{eq:normal_flow}
\end{equation}

where $\nabla \textit{I} =( I_x, I_y)$ is the intensity spatial gradient. Given the gradient direction as a unitary vector $\textbf{n} = (n_x,n_y)$, the normal flow speed amounts to
$u_n (\textbf{x}) = \textbf{n}(\textbf{x}) \cdot \textbf{u}(\textbf{x}) $
where $\cdot$ represents the inner product (see also \cite{fermuller_passive_1995}). Considering the normal flow instead of the optical flow, we obtain from Eq.~(\ref{eq:longuet}) by multiplying both sides with $\textbf{n} = (n_x,n_y)$ :
\begin{eqnarray}
u_n(\textbf{x})  = \frac{1}{Z(\textbf{x})} \textbf{n}(\textbf{x}) \cdot A(\textbf{x})\textbf{t} + \textbf{n}(\textbf{x}) \cdot B(\textbf{x})\textbf{w}
\label{eq:longuet_normal}
\end{eqnarray}



\subsection{Constraints for the objective function}
\label{sec:3.2}
The 3D motion parameters and the depth are estimated using various optimization constraints.
A first approximation minimizes the squared distances with respect to the model as in Eq.~(\ref{eq:squared_dist})
\begin{eqnarray}
	arg\,min_{~\textbf{t}, \textbf{w}} \sum_{i=1}^{N} \left\| \textbf{u}(\textbf{x}_{i}) - \frac{1}{Z_{i}} A(\textbf{x}_{i})\textbf{t} - B(\textbf{x}_{i})\textbf{w} \right\|_{2},
	\label{eq:squared_dist}
\end{eqnarray}

where $\textbf{x}_{i}$ is a position of the image $(x_{i},y_{i})$, and $N$ is the total number of flow estimates of the image. Thus, the system has $N$ constraints where we have N depth unknowns $Z_{i}$ and the  parameters for $\textbf{t}$ and $\textbf{w}$.

The classic approach removes the depth from Eq.~(\ref{eq:squared_dist}), obtaining the so-called \textit{epipolar constraint}.
It can be written in the form:

\begin{eqnarray}
	arg\,min_{~\textbf{t}, \textbf{w}} \sum_{i=1}^{N} \left\| \left( \textbf{u}(\textbf{x}_{i}) -  B(\textbf{x}_{i})\textbf{w} \right) \cdot \left( A(\textbf{x}_{i})\textbf{t} \right)^{\perp} \right\|_2,
	\label{eq:epipolar_weighted}
\end{eqnarray}

where $\left( A(\textbf{x}_{i})\textbf{t} \right)^{\perp}$ denotes the vector perpendicular to $A(\textbf{x}_{i})\textbf{t}$. This constraint is also referred in the literature simply as the bilinear constraint, as it is linear in the translation, or the rotation. Some works have demonstrated that optimizing this constraint introduces a statistical bias. This is avoided with a slightly modified version
\begin{eqnarray}
	arg\,min_{~\textbf{t}, \textbf{w}} \sum_{i=1}^{N} \left\| \left( \textbf{u}(\textbf{x}_{i}) -  B(\textbf{x}_{i})\textbf{w} \right) \cdot \frac{\left( A(\textbf{x}_{i})\textbf{t}\right)^{\perp}}{\left\| A(\textbf{x}_{i})\textbf{t} \right\|}  \right\|_2.
	\label{eq:epipolar_unweighted}
\end{eqnarray}

Many methods solve the bilinear contraints or weighted variations (e.g. Eq.~(\ref{eq:epipolar_unweighted})) in a 2-step approach: they first search for either the rotation or the translation and then solve for the other one. Other formulations, similar to the use of the essential matrix in the discrete case, solve first for intermediate parameters that encode both rotation and translation \cite{Kanatani1993}.

Another approach is to employ a scene model. In \cite{brodsky_structure_2000,ji_3dshape_2006} solutions were proposed that assume  3D planar piece-wise depth.
Let assume $P$ patches where $P<<N$; if all points $(x_{k},y_{k})$ in the image patch $p$ lie on the same plane $\Gamma_{p}$ then:
\begin{equation}
	\Gamma_{p} = \alpha_{p}~\frac{x_{k}}{f} + \beta_{p}~\frac{y_{k}}{f} + \gamma_{p}  = \frac{1}{Z_{k}},
	\label{eq:planar_eq}
\end{equation}

where $\textbf{v}_{p}=(\alpha_{p},\beta_{p},\gamma_{p})$ stands for the plane vector and $f$ is the focal length.

As mentioned before, in order to compute the 3D motion and relative depth, only the axis of the translational motion is required. The equation can be rewritten using $C_{k} =\left \| \textbf{t} \right \|  \left( Z_{k} \right)^{-1}$ to denote the scaled inverse depth, and $\tilde{\textbf{t}} = \textbf{t}/\left\|\textbf{t}\right\|$ to denote the translation axis. Due to this change, instead of optimizing for plane parameters $\textbf{v}_{p}$, we optimize for $\tilde{\textbf{v}}_{p}=\left\|\textbf{t}\right\|(\alpha_{p},\beta_{p},\gamma_{p})$. We then obtain an overdetermined system with $N$ equations for a total number of $3 * P + 5$ unknowns. Denoting individual patches as $p_i$, each with $K_{p_i}$ image motion measurements, we have:
\begin{eqnarray}
	arg\,min_{~\tilde{\textbf{t}}, \textbf{w}, \tilde{\textbf{v}}_{p_i}} \sum_{j=1}^{P} \sum_{k=1}^{K_{p_i}} \left\| \textbf{u}(\textbf{x}_{k}) - \Gamma_{p_i}(\tilde{\textbf{v}}_{p_i},\textbf{x}_{k}) A(\textbf{x}_{k})\tilde{\textbf{t}} - B(\textbf{x}_{k})\textbf{w} \right\|_{2}.
\label{eq:planar_constraint}
\end{eqnarray}

We can apply the same approach to normal flow. If our measurements are the normal flow speeds
$u_n(\textbf{x}_{k})= \textbf{n}(\textbf{x}_{k}) \cdot \textbf{u}(\textbf{x}_{k})$,
we obtain from equation \ref{eq:planar_constraint}
\begin{eqnarray}
	arg\,min_{~\tilde{\textbf{t}}, \textbf{w}, \tilde{\textbf{v}}_{p_i}} \sum_{j=1}^{P} \sum_{k=1}^{K_{p_i}} \left\| u_{n}(\textbf{x}_{k}) - \textbf{n}(\textbf{x}_{k}) \cdot \Gamma_{p_i}(\tilde{\textbf{v}}_{p_i},\textbf{x}_{k}) A(\textbf{x}_{k})\tilde{\textbf{t}} - \textbf{n}(\textbf{x}_{k}) \cdot B(\textbf{x}_{k})\textbf{w} \right\|_{2}.
\label{eq:planar_constraint_normalflow}
\end{eqnarray}
We have implemented for comparison a version of this method that uses regular patches to partition the image. One could first segment the scene to obtain the patches, which however as shown in \cite{ji_3dshape_2006} does not lead to significant accuracy improvement for the 3D motion estimation.

The above constraints require assumptions on the local smoothness of image motion, either for computing optical flow that is used in the epipolar constraint or indirectly in the assumption of planar scene patches. However, there is another weaker constraint, that can be used with normal flow only, thus avoiding these assumptions. This is
the so-called positive depth constraint. Since the scene in view is in front of the camera, its depth is positive, i.e. $Z > 0$.  From Eq.~(\ref{eq:longuet_normal}), subtracting the rotational component, we can derive that the derotated normal flow and the translational flow have to have same sign for $Z$ to be positive. Thus
\begin{eqnarray}
\left( u_{n}(\textbf{x})  - \textbf{n}(\textbf{x}) \cdot B(\textbf{x})\textbf{w} \right) \cdot \left( \textbf{n}(\textbf{x}) \cdot A(\textbf{x})\tilde{\textbf{t}} \right) > 0.
\label{eq:positive_depth_constraint}
\end{eqnarray}

A simple solution proposed in \cite{fermuller_observability_2000} optimizes this constraint with a voting function that counts the number of negative depth values \cite{fermuller_observability_2000}. One can search for the $\tilde{\textbf{t}}$ and $\textbf{w}$ with smallest number of normal flow values that make Eq.~(\ref{eq:positive_depth_constraint}) negative. This minimization can be expressed as
\begin{eqnarray}
	arg\,min_{~\tilde{\textbf{t}}, \textbf{w}} \sum_{i=1}^{N} \mathcal{V}(\textbf{x}_{i}, \tilde{\textbf{t}}, \textbf{w}) \quad \mbox{with}
	\label{eq:constraint_normal_original_minimization}
\end{eqnarray}
\begin{equation}
\begin{split}
\mathcal{V}(\textbf{x}, \tilde{\textbf{t}}, \textbf{w}) = \quad \quad \quad \quad \quad \quad \quad \quad \quad \quad \quad \quad \quad \quad \quad \quad \quad\\
\left\{\begin{array}{cl}
0 &\textrm{if}\left(u_{n}(\textbf{x}) -\textbf{n}(\textbf{x})\cdot B(\textbf{x})\textbf{w}\right)\cdot \left(\textbf{n}(\textbf{x})\cdot A(\textbf{x})\tilde{\textbf{t}} \right)>0 \\
1 &\textrm{if} \left( u_{n}(\textbf{x}) -\textbf{n}(\textbf{x})\cdot B(\textbf{x})\textbf{w}\right)\cdot \left(\textbf{n}(\textbf{x})\cdot A(\textbf{x})\tilde{\textbf{t}} \right)<0
\end{array}\right.
\end{split}
\label{eq:voting_original}
\end{equation}

A more robust solution considers only the sign of the normal flow \cite{fermuller_direct_1995}. By considering subsets of normal flow values, the axis of translation and axis of rotation can be constrained. However, the proposed constraint is weaker and if used by itself it can only provide bounds for the solution.
\begin{equation}
\begin{split}
\mathcal{V}_{r}(\textbf{x}, \tilde{\textbf{t}}, \textbf{w}) = \quad \quad \quad \quad \quad \quad \quad \quad \quad \quad \quad \quad \quad \quad \quad \quad \quad\\
\left\{ \begin{array}{cl}
1 & \textrm{if }u_{n}(\textbf{x})>0,\ \textbf{n}(\textbf{x}) \cdot B(\textbf{x})\textbf{w}<0,\ \textbf{n}(\textbf{x}) \cdot A(\textbf{x})\tilde{\textbf{t}}<0\\
1 & \textrm{if }u_{n}(\textbf{x})<0,\ \textbf{n}(\textbf{x}) \cdot B(\textbf{x})\textbf{w}>0,\ \textbf{n}(\textbf{x}) \cdot A(\textbf{x})\tilde{\textbf{t}}>0\\
0 & \textrm{otherwise}
\end{array}\right.
\end{split}
\label{eq:voting_robust}
\end{equation}
There are variations that ease the computation by removing either the rotational or translational motion from the previous constraint. The first variation considers that the rotational motion can be computed accurately using inertial sensors  (see \cite{corke_introduction_2007}). In this case, the rotational motion component is substracted from the normal flow speed, and one has to solve the minimization only for the translation. In practice, an additional step of sensor fusion to combine  the inertial and visual data  is required for this solution, because  the  sensors have different noise models and rates.

The second variation heavily simplifies the minimization. Given a translation, this constraint only considers the normal flow vectors orthogonal to the translational motion component. For these components the translational motion term is zero, i,e. $A(\textbf{x})\tilde{\textbf{t}} \cdot \textbf{n} = 0$. Algorithms that use these idea, search for the translational axis $\tilde{\textbf{t}}$, and for each axis solve an optimization in the rotational motion as
\begin{eqnarray}
	arg\,min_{~\textbf{w}} \sum_{j=1}^{M} \left\| u_n(\textbf{x}_{j})  - \textbf{n}(\textbf{x}_{j}) \cdot B(\textbf{x}_{j})\textbf{w}~\right\|,
	\label{eq:constraint_normal_perpendicular_transflow}
\end{eqnarray}

where $M$ is the number of points $\textbf{x}_{j}$ with normal flows orthogonal to the translation motion components and thus $M \leq N$. The main drawback here is that the number of points with translational flow perpendicular to the gradient can be much smaller than $N$, and thus the estimation becomes less accurate. Some works combined this with additional constraints to achieve a solution for the general motion case \cite{hui_determining_2015}.

\section{Our direct approach using normal flow}
\label{sec:4}
Let us denote the left hand side in Eq.~(\ref{eq:positive_depth_constraint}) as $f(\tilde{\textbf{t}},\textbf{w}, \textbf{x})$, i.e.
$f(\tilde{\textbf{t}},\textbf{w},\textbf{x}) = \left( u_{n}(\textbf{x})  - \textbf{n}(\textbf{x}) \cdot B(\textbf{x})\textbf{w} \right) \cdot \left( \textbf{n}(\textbf{x}) \cdot A(\textbf{x})\tilde{\textbf{t}} \right)$.
We then model the inequality in Eq.~(\ref{eq:positive_depth_constraint}) using the negative ReLu function, which we denote as $\mathcal{H}$, and
reformulate the optimization in Eq.~(\ref{eq:constraint_normal_original_minimization}) as
\begin{eqnarray}
	arg\,min_{~\tilde{\textbf{t}}, \textbf{w}} \sum_{i=1}^{N} \mathcal{H}(f(\tilde{\textbf{t}}, \textbf{w}, \textbf{x}_{i})
	\label{eq:optimization}
\end{eqnarray}

where
\begin{equation}
\mathcal{H}(x) = \left\{ \begin{array}{cl}
-x & \textrm{if }x\leq0\\
0 & \textrm{otherwise}
\end{array}\right.
\label{eq:heaviside_convex_approx}
\end{equation}

However $\mathcal{H}(x)$ is not strong convex since $\mathcal{H}''(x) = 0$, and thus it cannot be used in  Newton-type optimization methods. To use it with an interior point method, we only need the objective function and its gradient.
Since the gradient $\mathcal{H}'(x)$ has a singular point at $0$, we use the following smooth approximation, as is common in machine learning:
\begin{equation}
\mathcal{H}'(x) = \left\{ \begin{array}{cl}
-1 & \textrm{if }x\leq-\epsilon\\
-1/2 & \textrm{if }-\epsilon<x<\epsilon\\
0 & \textrm{otherwise}
\end{array}\right.
\label{eq:heaviside_convex_gradient}
\end{equation}

To solve for the 3D motion, we iteratively solve for $\tilde{\textbf{t}}$ and $\textbf{w}$. The method starts by searching for the translation axis. Given a candidate translation, we optimize the objective function
$\sum_{i=1}^{N} \mathcal{H}(f(\tilde{\textbf{t}}, \textbf{w}, \textbf{x}_{i}))$
 to solve for the rotation $\textbf{w}$ providing the gradient, which in this case is
\begin{equation}
\frac{\partial \sum_{i=1}^{N}\mathcal{H}(f(\tilde{\textbf{t}}, \textbf{w}, \textbf{x}_{i}))}{\partial \textbf{w}} = \sum_{i=1}^{N}\mathcal{H}'(f(\tilde{\textbf{t}}, \textbf{w},\textbf{x}_{i}) \cdot f'(\tilde{\textbf{t}}, \textbf{w},\textbf{x}_{i})
\label{eq:gradient_H}
\end{equation}
with
\begin{equation}
\begin{split}
f'(\tilde{\textbf{t}}, \textbf{w},\textbf{x}) = \left( \textbf{n}(\textbf{x})
\begin{bmatrix}
    -f & 0 & x \\
    0 & -f & y
\end{bmatrix}
\tilde{\textbf{t}} \right)
\cdot \quad \quad \quad \quad \\
\left(
\begin{bmatrix}
    \frac{x y}{f} & \left(\frac{y^2}{f} + f\right) \\
    \left(-\frac{x^2}{f}-f\right) & -\frac{x y}{f} \\
     y & -x
\end{bmatrix} \cdot
\left(-\textbf{n}(\textbf{x})^{T} \right) \right).
\end{split}
\label{eq:positive_depth_constraint_gradient}
\end{equation}
After having solved for the rotation, the translation is re-estimated using
Eq.~(\ref{eq:optimization}))
to obtain a more accurate solution for the translation axis.
\begin{figure}[tb]
\begin{center}
\begin{minipage}[b]{0.31\textwidth}
	\centering
 	 \includegraphics[width=\textwidth]{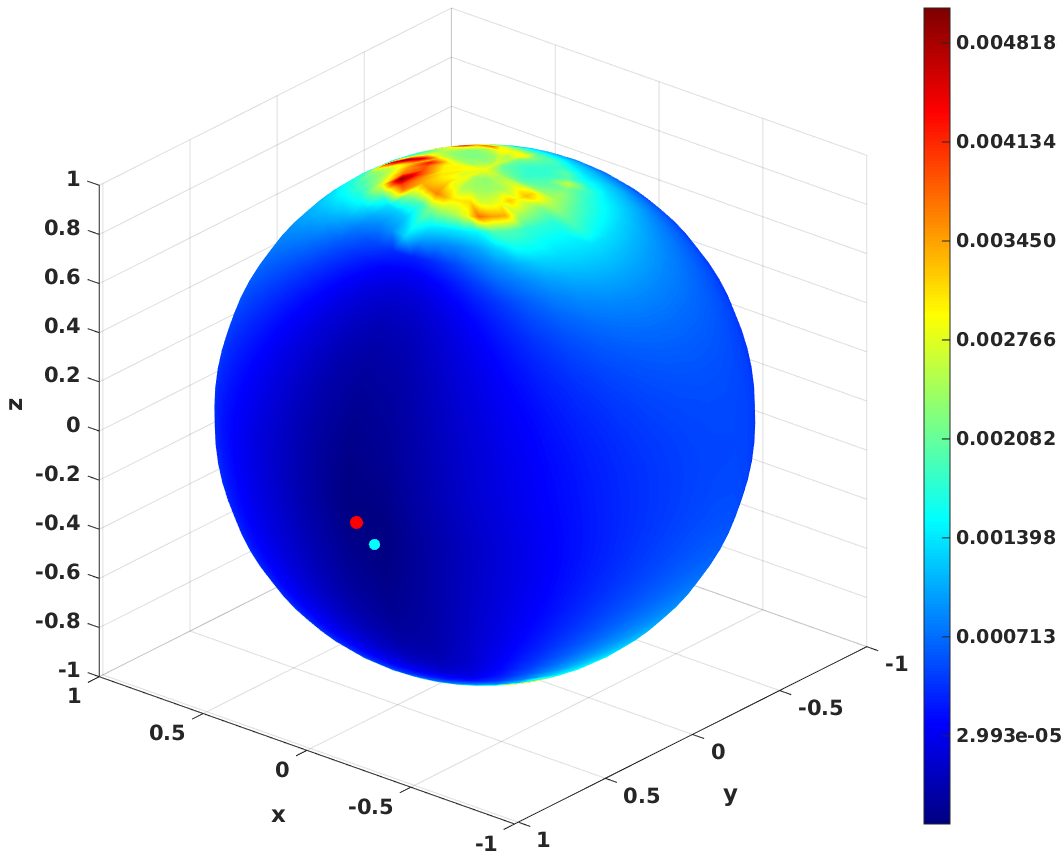}
\end{minipage}
\hspace{0.05cm}
\begin{minipage}[b]{0.31\textwidth}
	\centering
 	 \includegraphics[width=\textwidth]{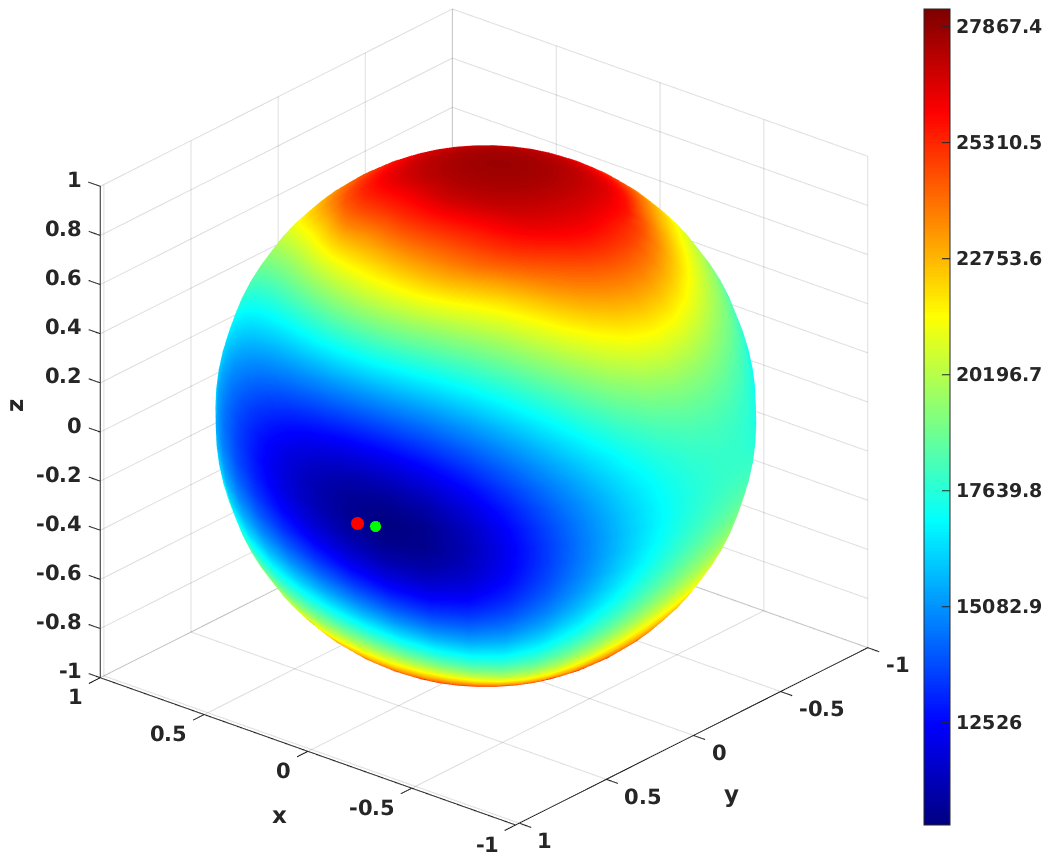}
\end{minipage}
\hspace{0.05cm}
\begin{minipage}[b]{0.31\textwidth}
	\centering
 	 \includegraphics[width=\textwidth]{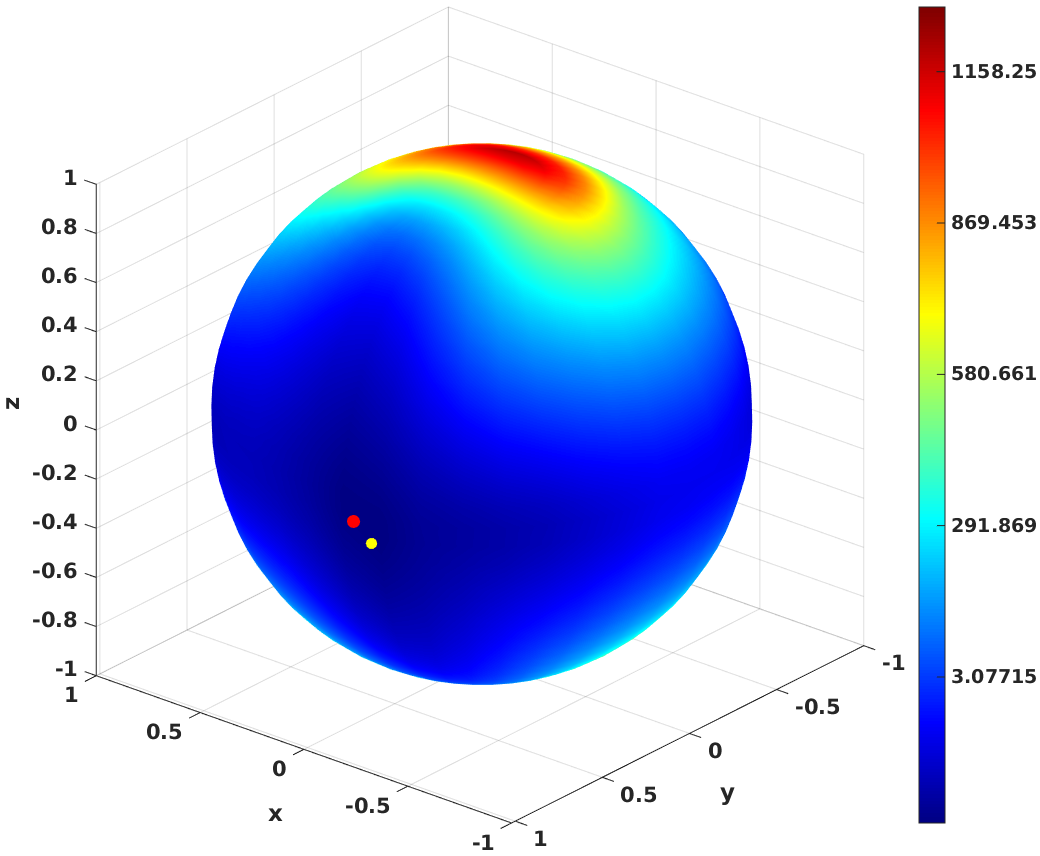}
\end{minipage}
\end{center}
\caption{Illustrative example of residual error surfaces for a randomly generated 3D motion, shown in the 2D subspace mapped onto a sphere surface that represents the translation direction. From left to right, top row: a) bilinear constraint from Eq.~(\ref{eq:epipolar_unweighted}), b) epipolar constraint assuming planar patches in Eq.~(\ref{eq:planar_constraint}), c) positive depth constraint with normal flow vectors in Eq.~(\ref{eq:positive_depth_constraint}). The ground-truth is marked with a red dot and  the estimated solution  with a colored smaller dot.}
\label{fig:error_surface_constraints}
\end{figure}

We visualize the behavior of the different minimization functions.
Fig.~\ref{fig:error_surface_constraints} shows the residuals of the optimization in the 2D subspace of translation directions on the surface of the sphere. For this illustrative example, we chose a random 3D motion and randomly generated depth values. For each possible translation, the residual is computed by solving for the optimal rotation. The ground-truth is marked with a red dot and the results from the different optimizations are marked with smaller dots in different colors.

Sphere \textit{a} shows the optimization for the bilinear constraint in Eq.~(\ref{eq:epipolar_unweighted}). Sphere \textit{b} shows  the optimization of the constraint assuming planar patches as in Eq.~(\ref{eq:planar_constraint}). Finally, sphere \textit{c} shows the results for the optimization using the positive depth constraint as in Eq.~(\ref{eq:positive_depth_constraint}).

Sphere \textit{b} shows a smaller region of low-valued residuals compared to the other two, but the optimization is done for the motion and planar shape parameters of all patches used. A small number of patches violates the initial hypothesis that assumes all points in the patch to lie on the same plane, and more patches increase the number of unknowns, increasing also the complexity of the minimization. The surfaces due to the the epipolar constraint in \textit{a} and the depth positive constraint \textit{c} show a similar minimum error region, although sphere \textit{c} has a smoother surface, and in this case, a smaller error between the computed solution and the ground-truth. However, for all the residuals shown here, the optical and normal flow values are the  ground-truth, while in real applications there is noise.

\subsection{Refining positive depth constraint}
\label{subsec:4.1}
In order to refine the solution detailed in \S\ref{sec:4}, a second step is added estimating first the 3D structure as follows
\begin{eqnarray}
C(\textbf{x}) = Z(\textbf{x}) / \left\| \textbf{t} \right\|= \frac{ A(\textbf{x})\hat{\textbf{t}} \cdot \textbf{n}(\textbf{x}) }{  u_{n}(\textbf{x})  - B(\textbf{x})\hat{\textbf{w}} \cdot \textbf{n}(\textbf{x}) }
\label{eq:structure_from_normalflow}
\end{eqnarray}
where $C$ stands for the structure and represents the depth up to a factor. Here $\hat{\textbf{t}}$ represents the translation axis estimated with the previous method. After that, a regularization process is performed on the structure using the inpainting method in \cite{janoch_category_2013}. This method has been designed for reconstructing parts of depth maps that are lost e.g. when using infrared-based Kinect sensors. In our case, the estimation is sparse since normal flow is only computed at locations of large image gradients such as at edges or object contours. The inpainting method reconstructs the depth at smooth surfaces while regularizing the estimates at object contours. It uses a second-order smoothness assumption as is common in natural images (see \cite{grimson_images_1981}). However, this assumption is violated close to object contours due to the depth discontinuities.

After the the depth regularization, a simpler least-squared minimization can be performed to obtain refined motion estimates as
\begin{eqnarray}
	arg\,min_{~\tilde{\textbf{t}}, \textbf{w}} \sum_{i=1}^{N} \left\| u_{n}(\textbf{x}_{i}) -  \left( C(\textbf{x}_{i}) A(\textbf{x}_{i})\tilde{\textbf{t}} - B(\textbf{x}_{i})\textbf{w} \right) \cdot \textbf{n}(\textbf{x}_{i}) \right\|_2.
	\label{eq:lsminimization_structure}
\end{eqnarray}

In our approach, these two steps are iteratively repeated further refining the solutions for the 3D motion and the 3D structure.  The stop criterion is designed with convergence threshold based on the difference between consecutive depth estimates.

\section{Experiments}
\label{sec:5}
This section presents: 1) an evaluation of our direct method using different metrics on various datasets, 2) a comparison of our direct approach with methods that use optical flow and with other direct methods that use normal flow, and 3) an evaluation of our method's recovery of 3D structure.

We evaluate the ego-motion estimation on the following four datasets:
\begin{enumerate}
\item The Artificial dataset contains 5000 random sequences of images of size 150~x~150 with a field of view (FOV) of 30$^{\circ}$, and an image plane of dimension 0.01~x~0.01 meters. The rotational velocity is up to 20$^{\circ}$ per frame, the translational velocity is up to 3 meters per frame, and the maximum virtual depth of the scene is 10 meters.

\item The Yosemite sequence \cite{YOS_barron_2017} has images of size  316~x~252, and is a simulation of a fly-through the Yosemite Valley with divergent motion and translational motion in the clouds. However, the clouds, as is common, have been masked out. The translation is $\textbf{t}=[0,0.17,0.98]^{T}$ with a speed of 34.8 pixels per frame, the rotational motion is $\textbf{w}=[0.0133, 0.0931, 0.0162]^{T}$ in degrees per frame, and the FOV is 40$^{\circ}$.

\item The Fountain sequence \cite{FOU_raudies_2017} is a synthetic sequences with images of size  320~x~240 of a curvilinear motion featuring a patio sequence surrounded by columns and a central fountain. The ground-truth optical flow, 3D pose, 3D velocity, and depth are provided. The translation is $\textbf{t}=[-0.2578,0.0872,0.9622]^{T}$ with a speed of 2.5 pixels per frame, the rotational motion is $\textbf{w}=[-0.125, 0.20, -0.125]^{T}$ degrees per frame, and the FOV is 40$^{\circ}$.

\item The Kitti dataset \cite{geiger_ready_2012} for visual odometry is a popular set of 22 driving sequences of stereo road scenes. These are long natural scenes with different trajectory directions and speeds. To the best of our knowledge, our study is the first  to evaluate a direct 3D motion estimation method on such a complex real-world dataset. We have used 11 of the sequences for which the ground-truth is provided: the total number of frames is more than 22500. Since there is very large inter-frame displacement  up to 50 pixels, making it impossible to estimate accurate intensity gradient, we interpolate frames to reduce the maximum displacement to 3 to 5 pixels.
\end{enumerate}

For the evaluation of 3D motion estimation we use two error metrics: the average angular error (AAE) which is defined as the average angular distance between the ground-truth $\textbf{u}$ and the estimated motion vector $\hat{\textbf{u}}$ as
\begin{equation}
AAE = \frac{1}{N}\sum_{i=1}^{N} \arccos \left( \frac{\hat{\textbf{u}}(\textbf{x}_{i})^{t}\textbf{u}(\textbf{x}_{i})}{\left\| \hat{\textbf{u}}(\textbf{x}_{i}) \right\| \left\| \textbf{u}(\textbf{x}_{i})\right\|} \right),
\label{eq:ae}
\end{equation}

 and the EPE, which is the average euclidean distance between the ground-truth and the estimated motion vector, defined as
\begin{equation}
EPE = \frac{1}{N}\sum_{i=1}^{N} \left\| \textbf{u}(\textbf{x}_{i})-\hat{\textbf{u}}(\textbf{x}_{i}) \right\|.
\label{eq:epe}
\end{equation}

Since we are estimating only the translation direction, we evaluate the AAE and EPE for rotation, but only the EPE for translation. For the Kitti dataset we do not provide the rotational AAE, because the rotational direction varies largely making it difficult to interpret this error.

We also use two error metrics to evaluate the estimated structure: the mean absolute error (MAE),
defined as the average absolute difference between the estimated 3D structure (times the translational speed) $\hat{d}$ and the disparity ground-truth $d$
\begin{equation}
MAE = \frac{1}{N}\sum_{i=1}^{N} \lvert d(\textbf{x}_{i})-\hat{d}(\textbf{x}_{i}) \rvert,
\label{eq:mae}
\end{equation}
 and the PoBP,  defined as the percentage of points with a MAE greater than 1 (see \cite{scharstein_taxonomy_2002}). This error gives a measure of the accuracy of the recovered structure at object contours.

\subsection{Our direct approach for 3D motion estimation}
\label{subsec:5.1}
\begin{figure}[tb]
\begin{center}	
	\begin{minipage}[b]{0.22\textwidth}
		\centering
 		 \includegraphics[width=\textwidth]{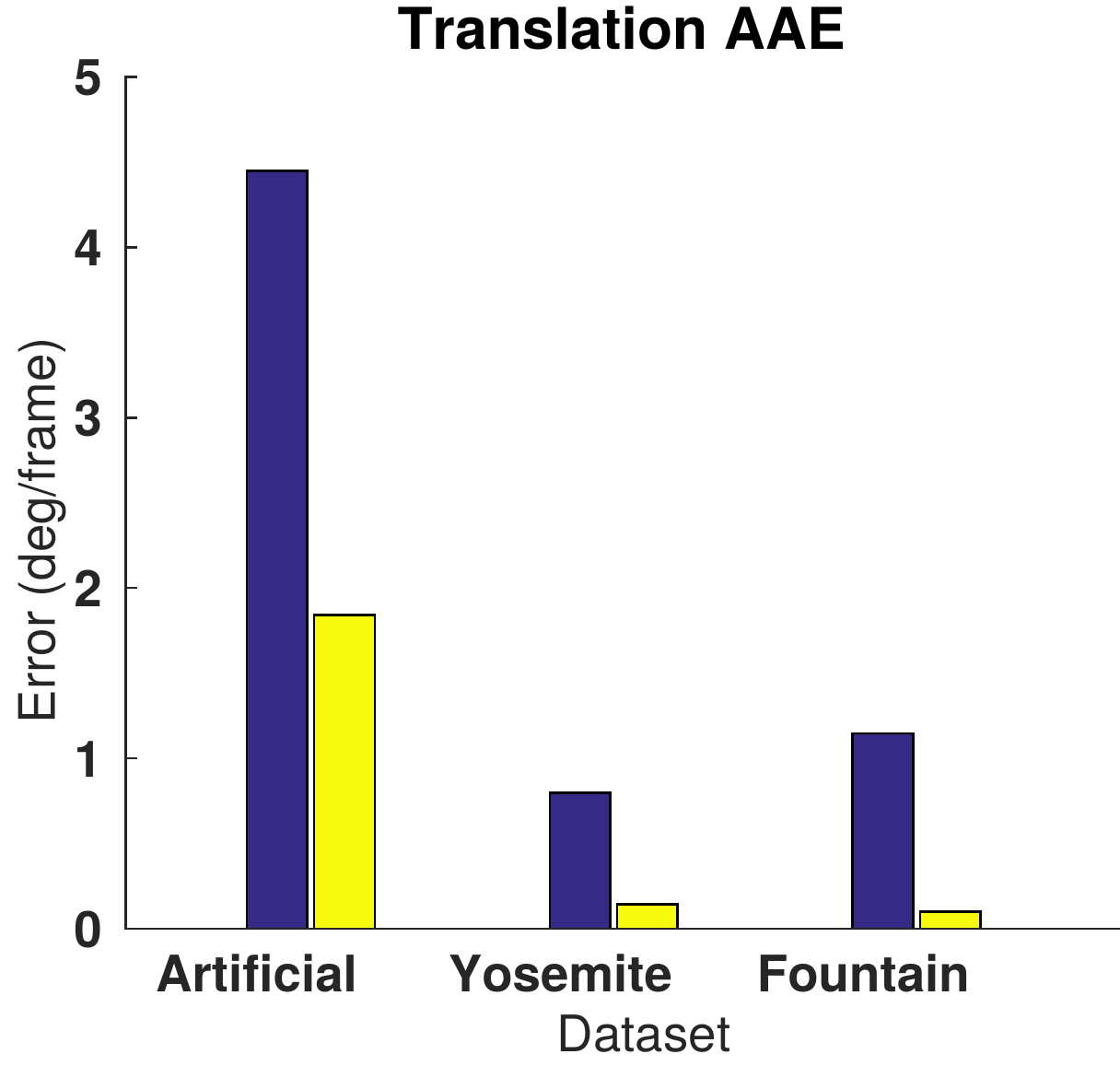}
	\end{minipage}
	\hspace{0.7cm}
	\begin{minipage}[b]{0.22\textwidth}
		\centering
	 	 \includegraphics[width=\textwidth]{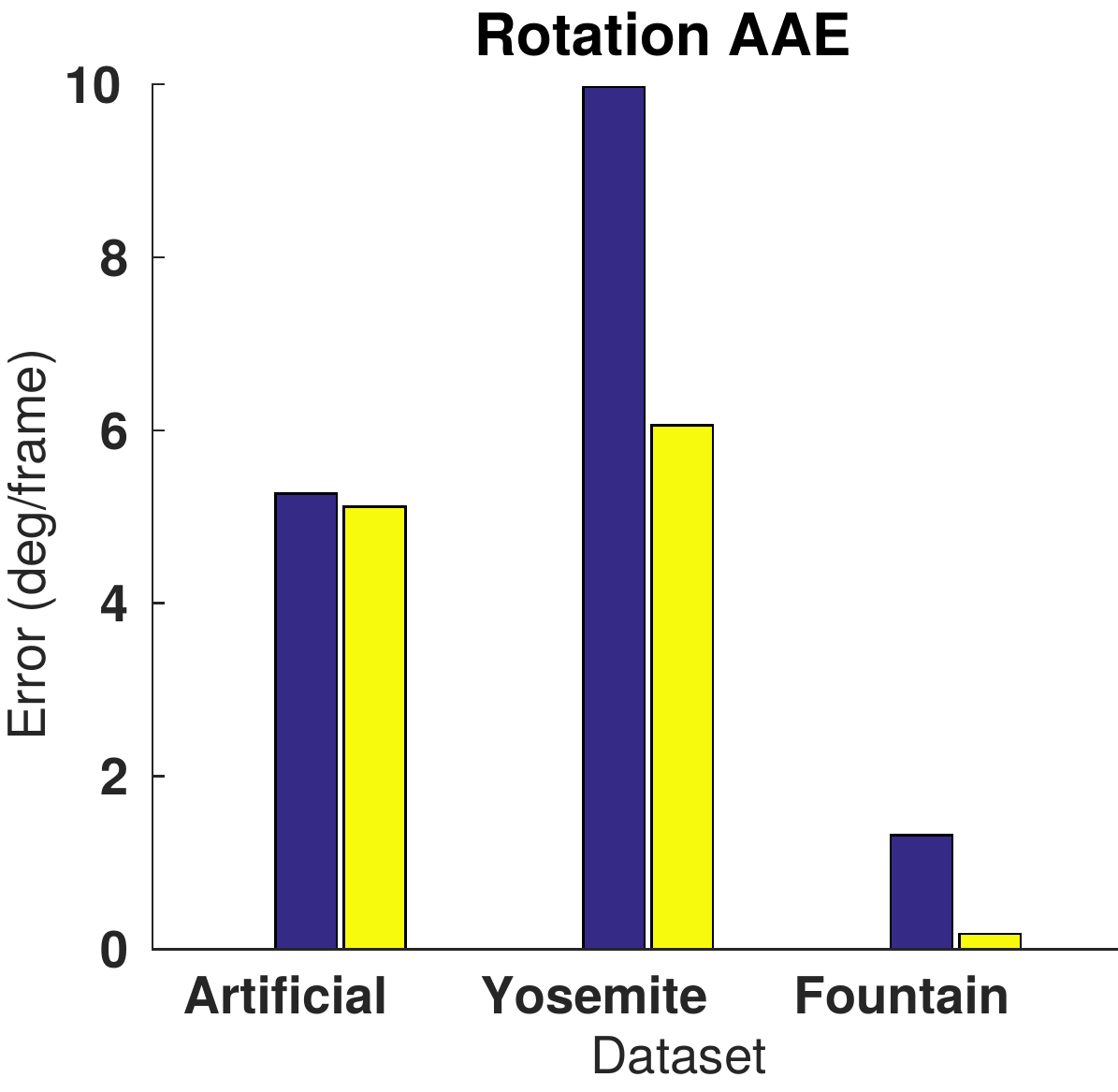}
	\end{minipage}
	\hspace{0.7cm}
	\begin{minipage}[b]{0.22\textwidth}
		\centering
 		 \includegraphics[width=\textwidth]{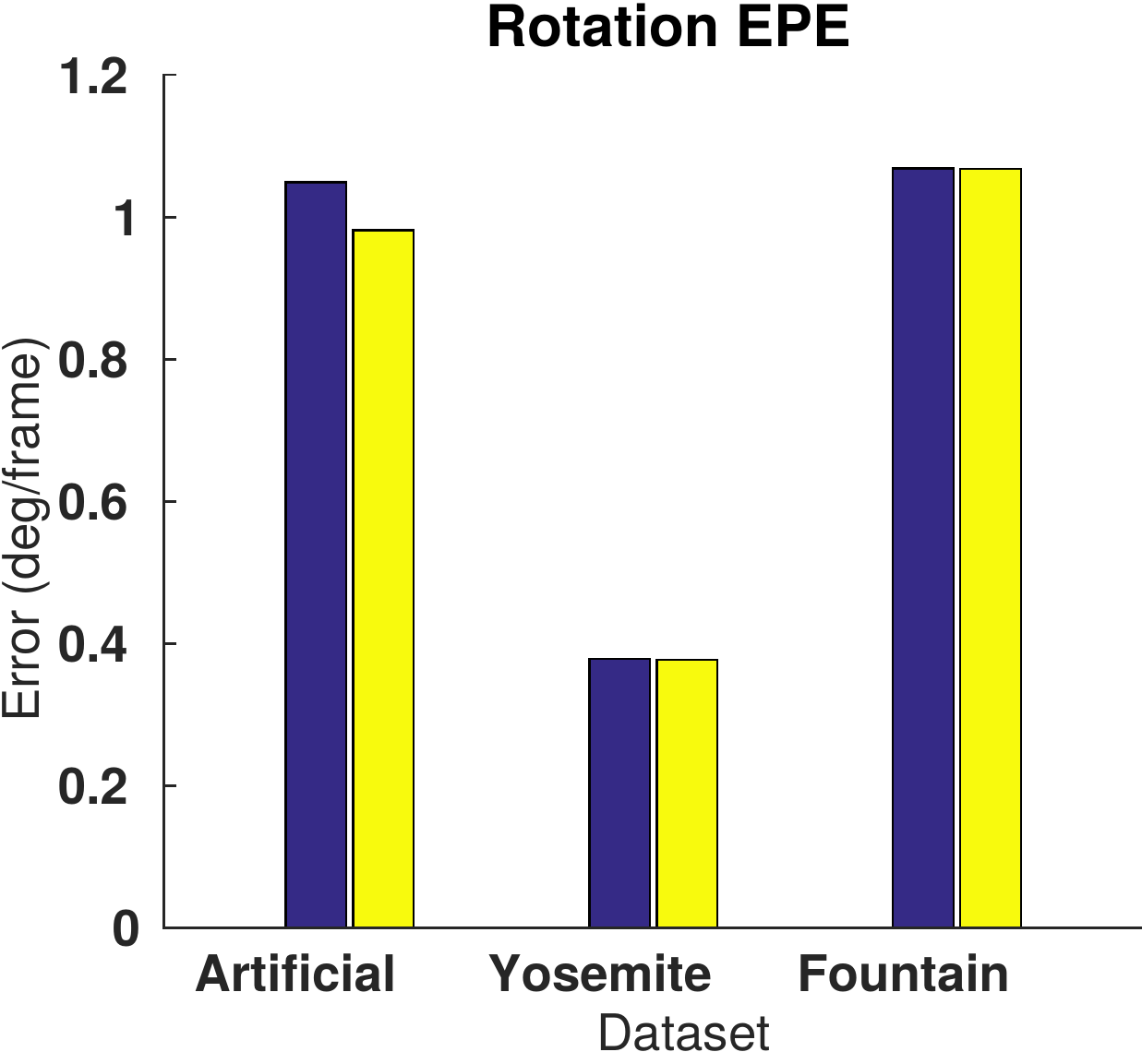}
	\end{minipage}
	\hspace{3.5cm}	
	\begin{minipage}[b]{0.15\textwidth}
		\centering
	\end{minipage}
	
%

	\vspace{0.01cm}	
	
	\begin{minipage}[b]{0.22\textwidth}
		\centering
 		 \includegraphics[width=\textwidth]{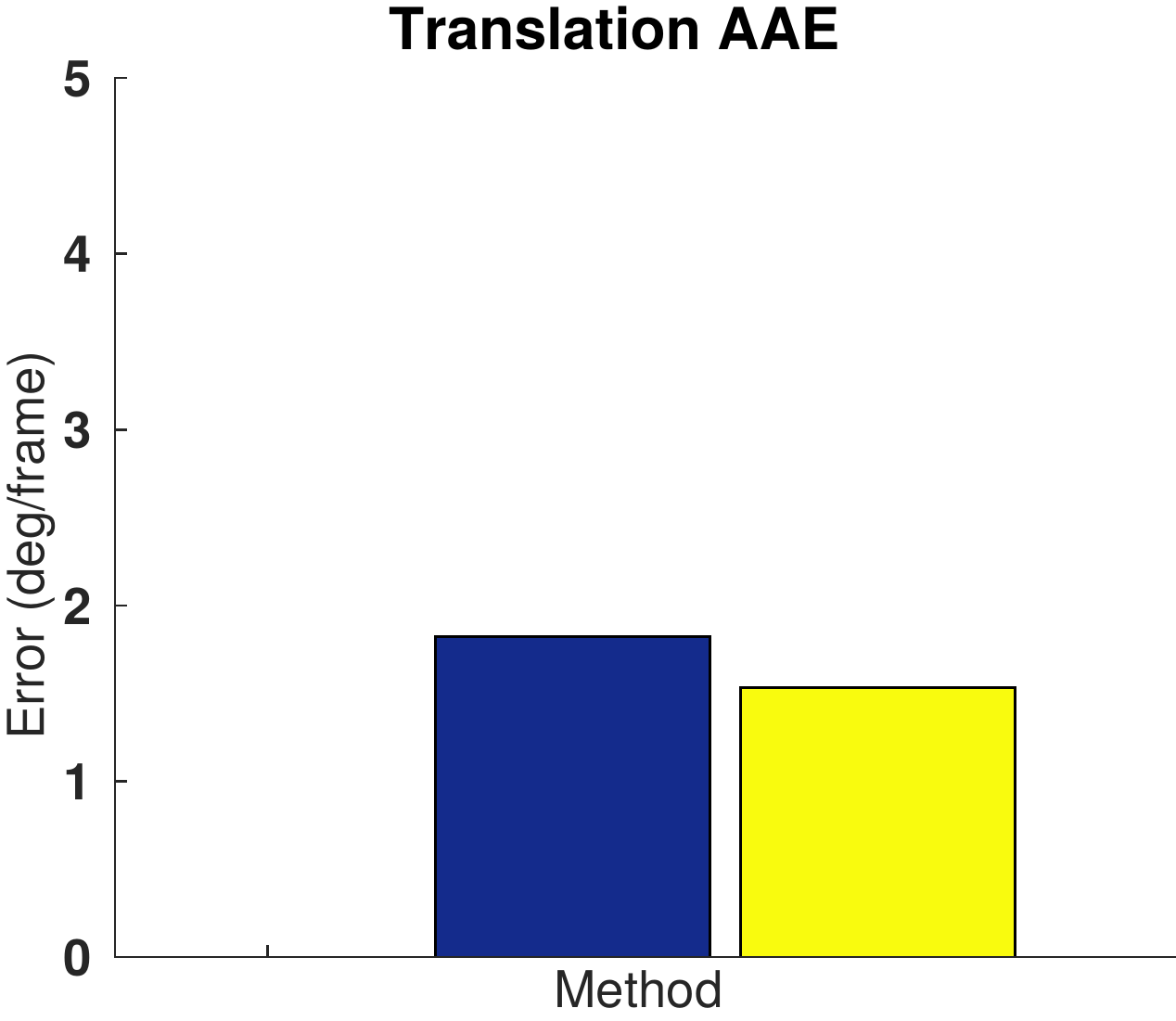}
	\end{minipage}
	\hspace{0.7cm}
	\begin{minipage}[b]{0.22\textwidth}
		\centering
	 	 \includegraphics[width=\textwidth]{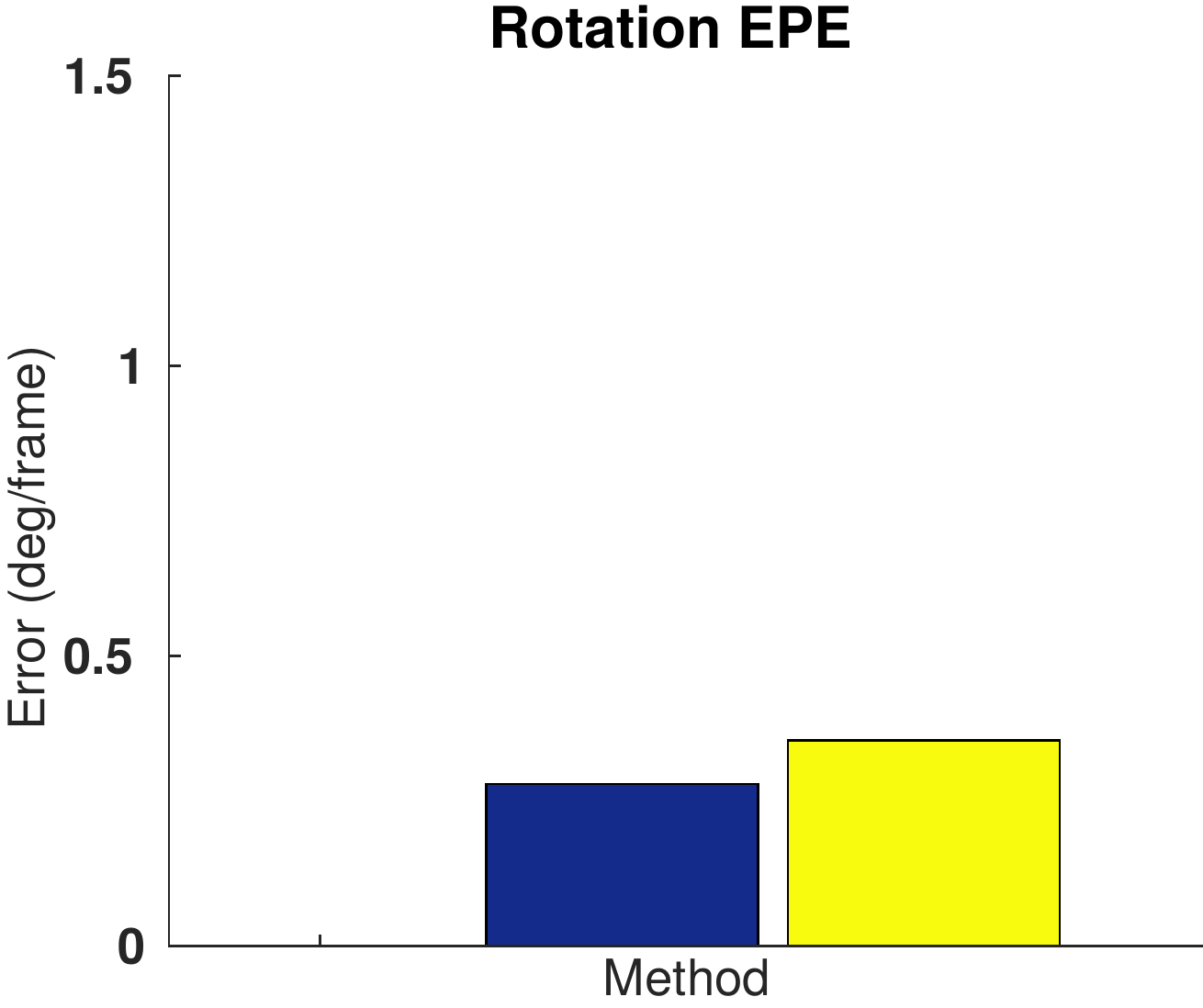}
	\end{minipage}
	\hspace{0.7cm}
	\begin{minipage}[b]{0.22\textwidth}
		\centering
	\end{minipage}	
	\begin{minipage}[b]{0.2\textwidth}
		\centering
		\includegraphics[width=\textwidth]{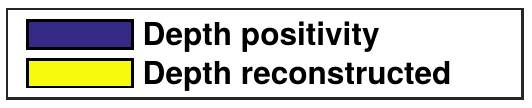}
	\end{minipage}
\end{center}
\caption{Error plots for 3D motion estimation estimated using our proposed method. The first row shows the translation AAE and rotation AAE and EPE for the Artificial, Yosemite, and Fountain datasets. The second row shows the translation AAE and the rotation EPE for the Kitti dataset, averaging over all frames and  sequences. For the artificial dataset, the normal flow is estimated in a random direction using the ground-truth; for the other datasets, the normal flow is computed using the spatio-temporal gradients. The blue and yellow bars show the error when using only the refined positive depth constraint, and after normalizing with the 3D structure reconstruction.}
\label{fig:nflow_baseline}
\end{figure}

The plots in Fig.~\ref{fig:nflow_baseline} illustrate the results of 3D motion estimation discussed in Section \S\ref{sec:4} for
a) the depth-independent minimization using the depth positivity constraint, and b) after inclusion of the refinement step based on the depth map. The first row shows the translational AAE and the rotational AAE and EPE for the Artificial, Yosemite, and Fountain datasets. After the 3D reconstruction, the error is reduced substantially for all datasets. For the Kitti dataset, the average error reduction is 10\% to 15\% for the translational AAE. The rotational EPE increases, but this increase is negligible for the given driving scenario, where the rotation is very small.


\begin{figure}[tb]
\begin{center}
\begin{minipage}[b]{0.25\textwidth}
	\centering
 	\includegraphics[width=\textwidth]{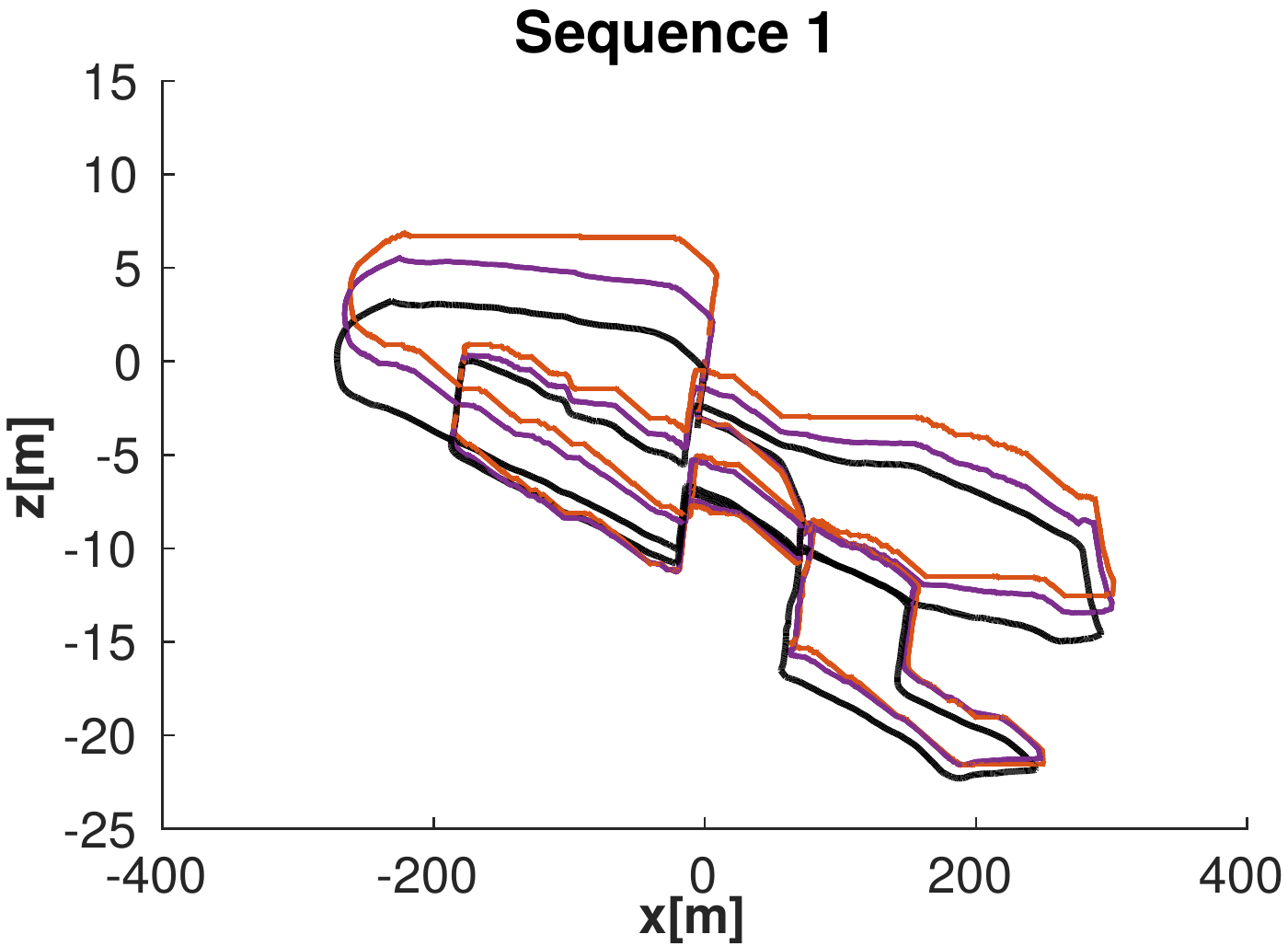}
\end{minipage}
\hspace{0.2cm}
\begin{minipage}[b]{0.25\textwidth}
	\centering
 	\includegraphics[width=\textwidth]{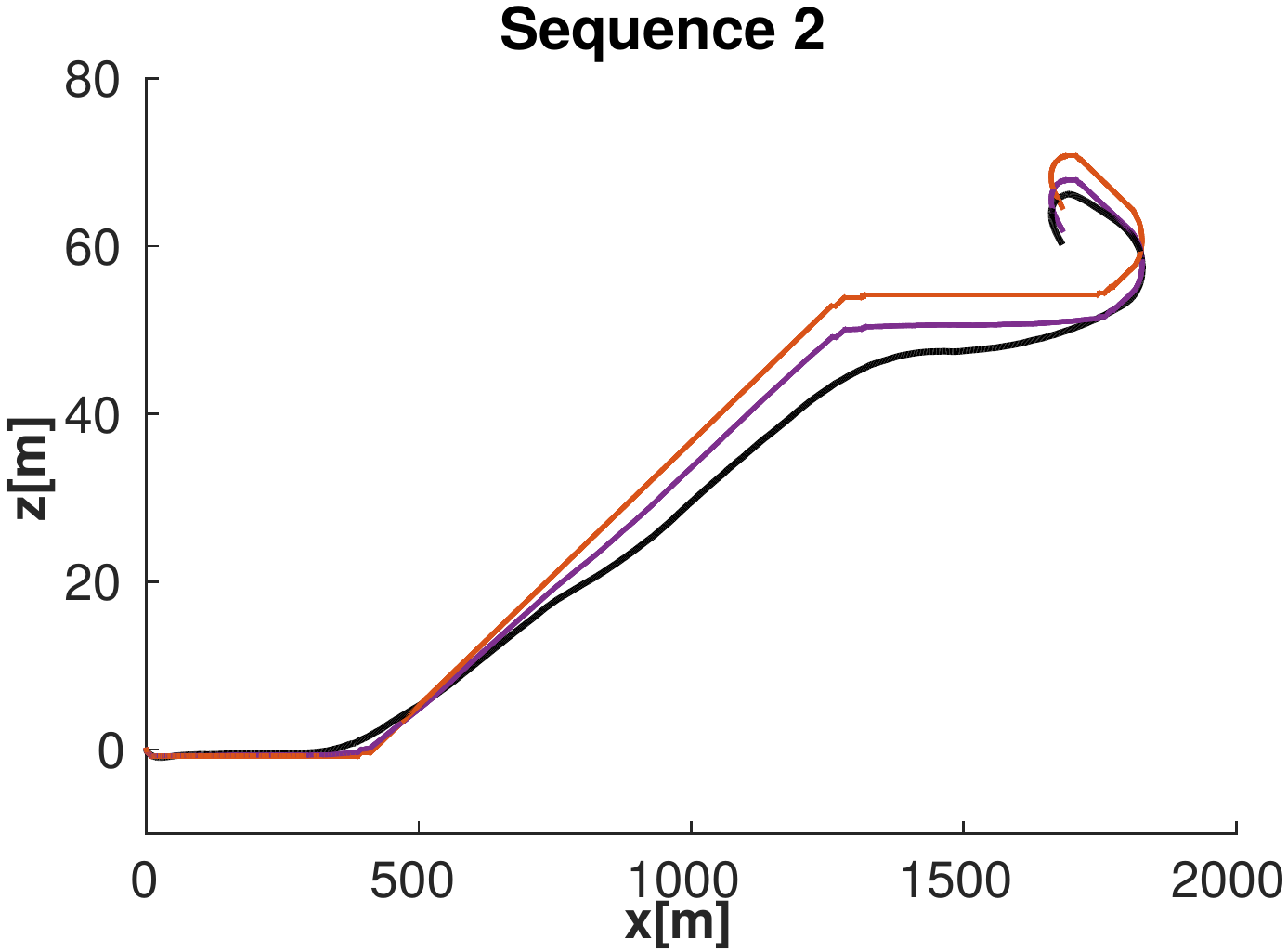}
\end{minipage}
\hspace{0.2cm}
\begin{minipage}[b]{0.25\textwidth}
	\centering
 	\includegraphics[width=\textwidth]{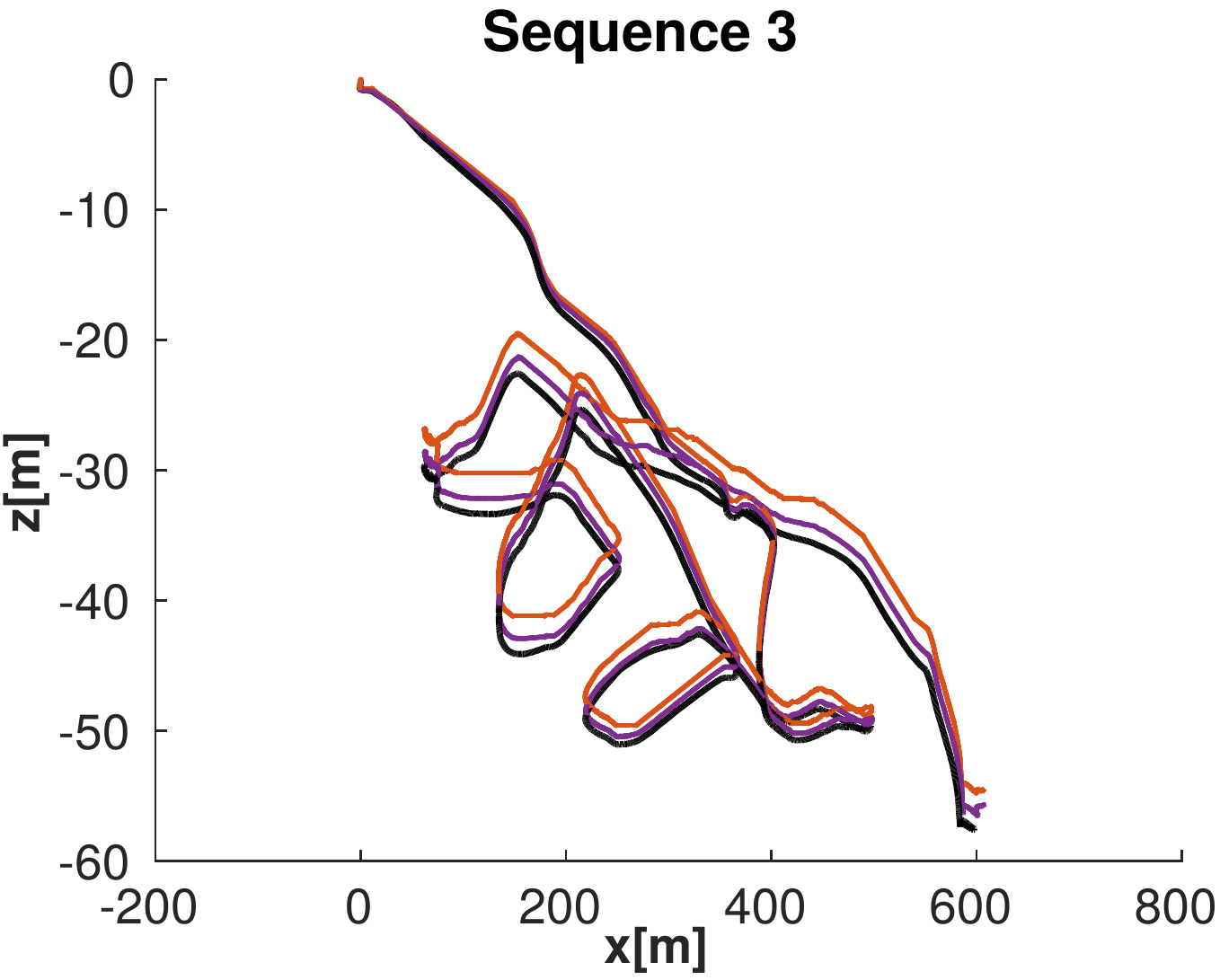}
\end{minipage}

\vspace{0.05cm}

\begin{minipage}[b]{0.25\textwidth}
	\centering
 	\includegraphics[width=\textwidth]{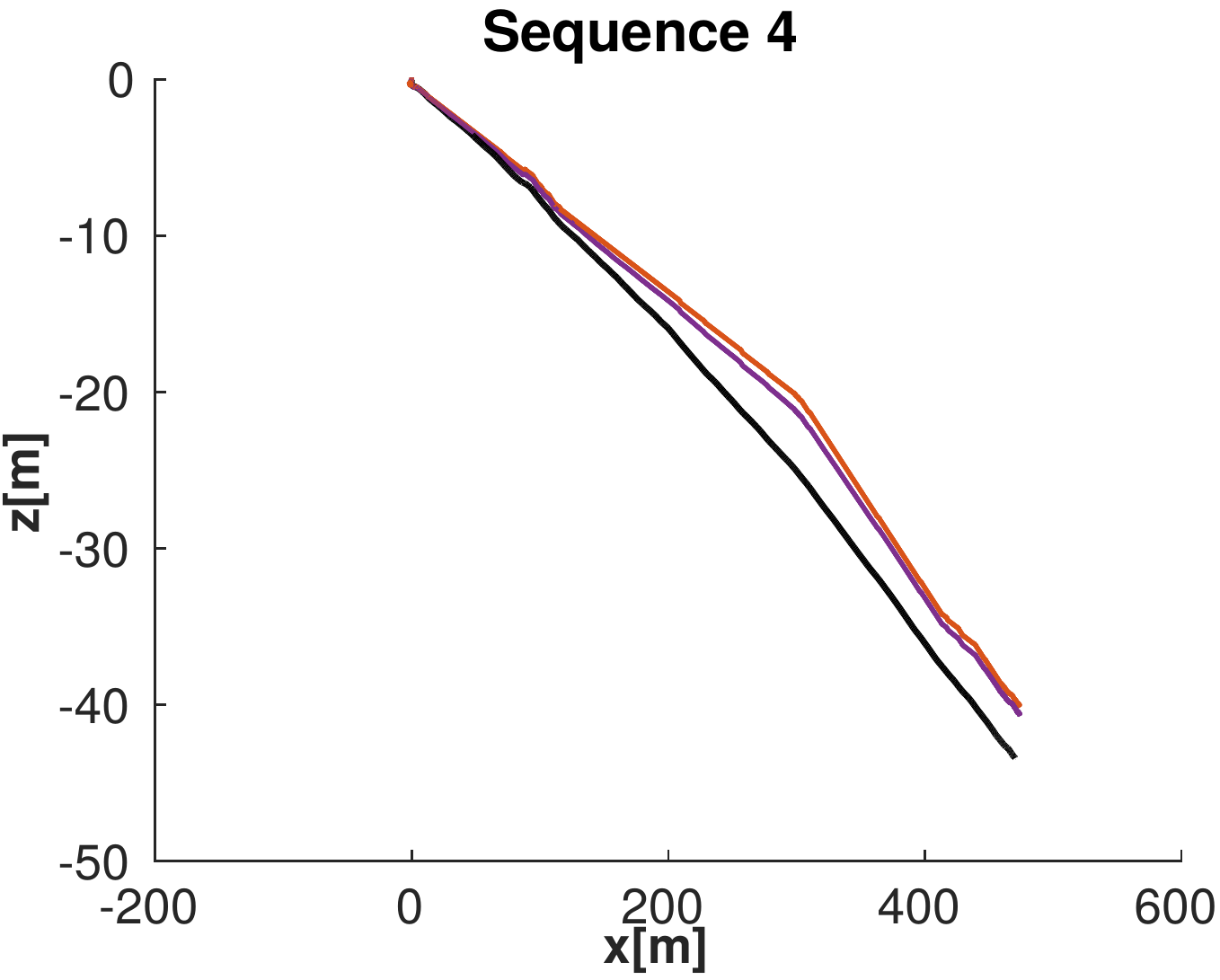}
\end{minipage}
\hspace{0.2cm}
\begin{minipage}[b]{0.25\textwidth}
	\centering
 	\includegraphics[width=\textwidth]{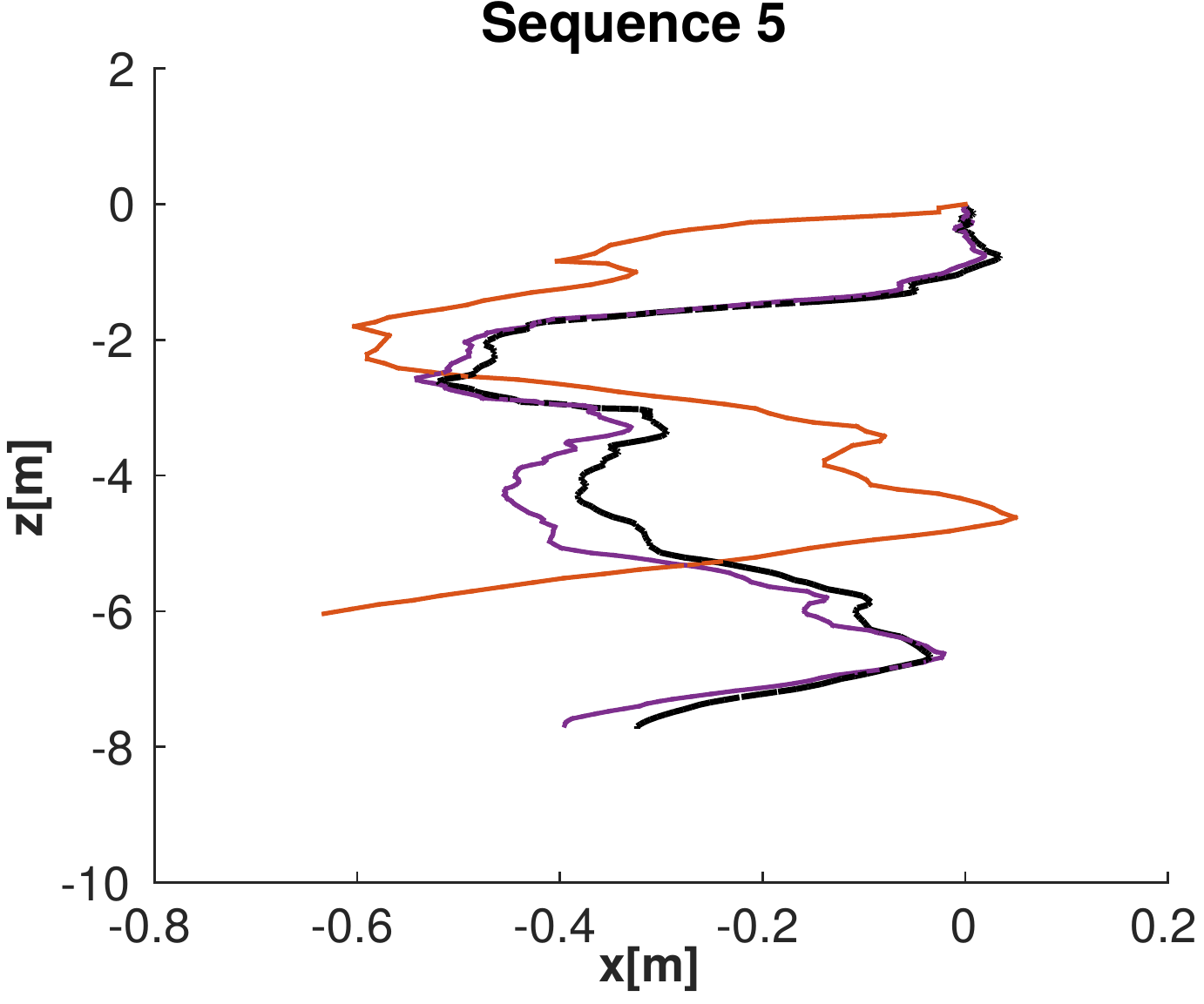}
\end{minipage}
\hspace{0.2cm}
\begin{minipage}[b]{0.25\textwidth}
	\centering
 	\includegraphics[width=\textwidth]{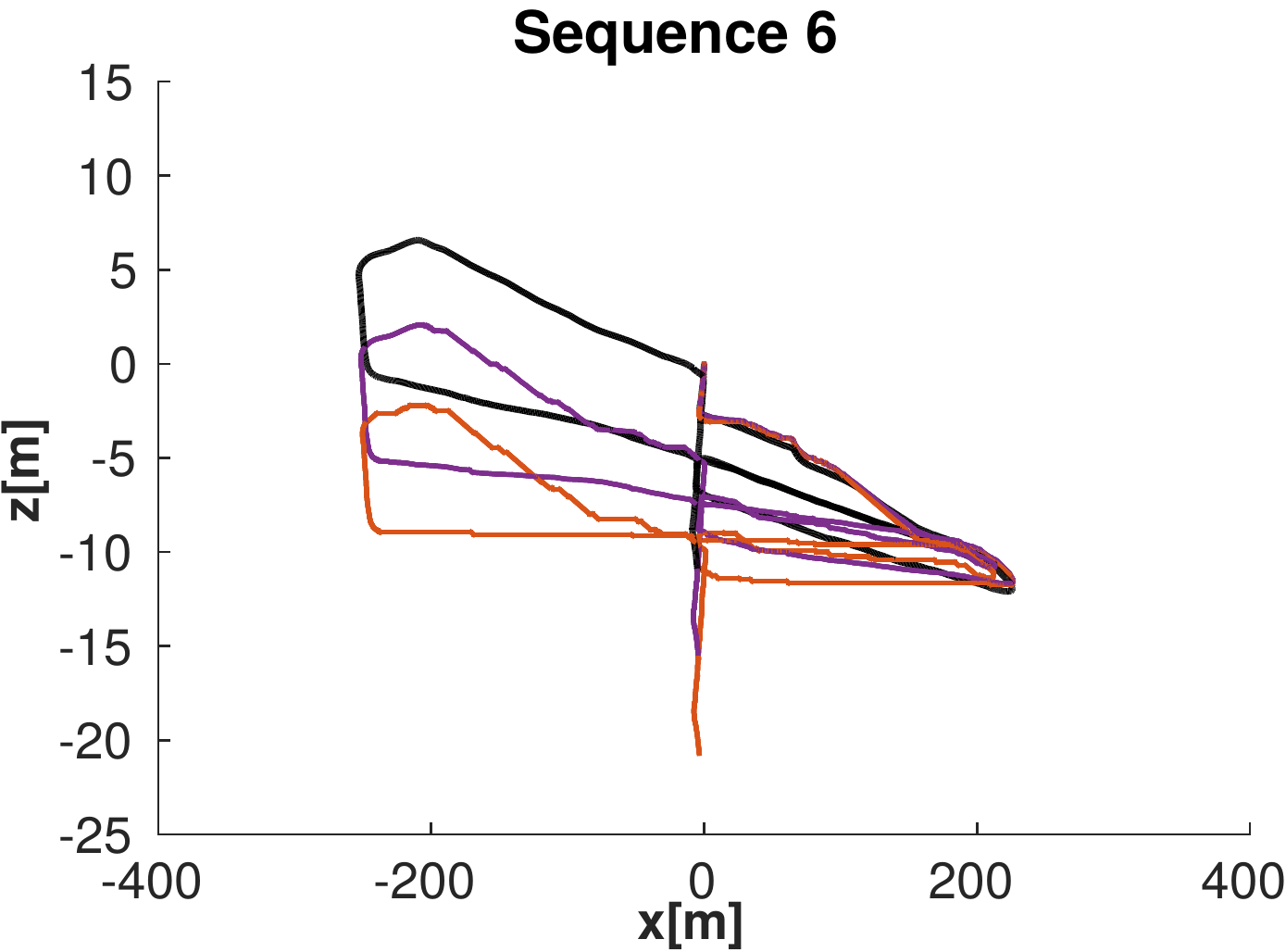}
\end{minipage}

\vspace{0.05cm}

\begin{minipage}[b]{0.25\textwidth}
	\centering
 	\includegraphics[width=\textwidth]{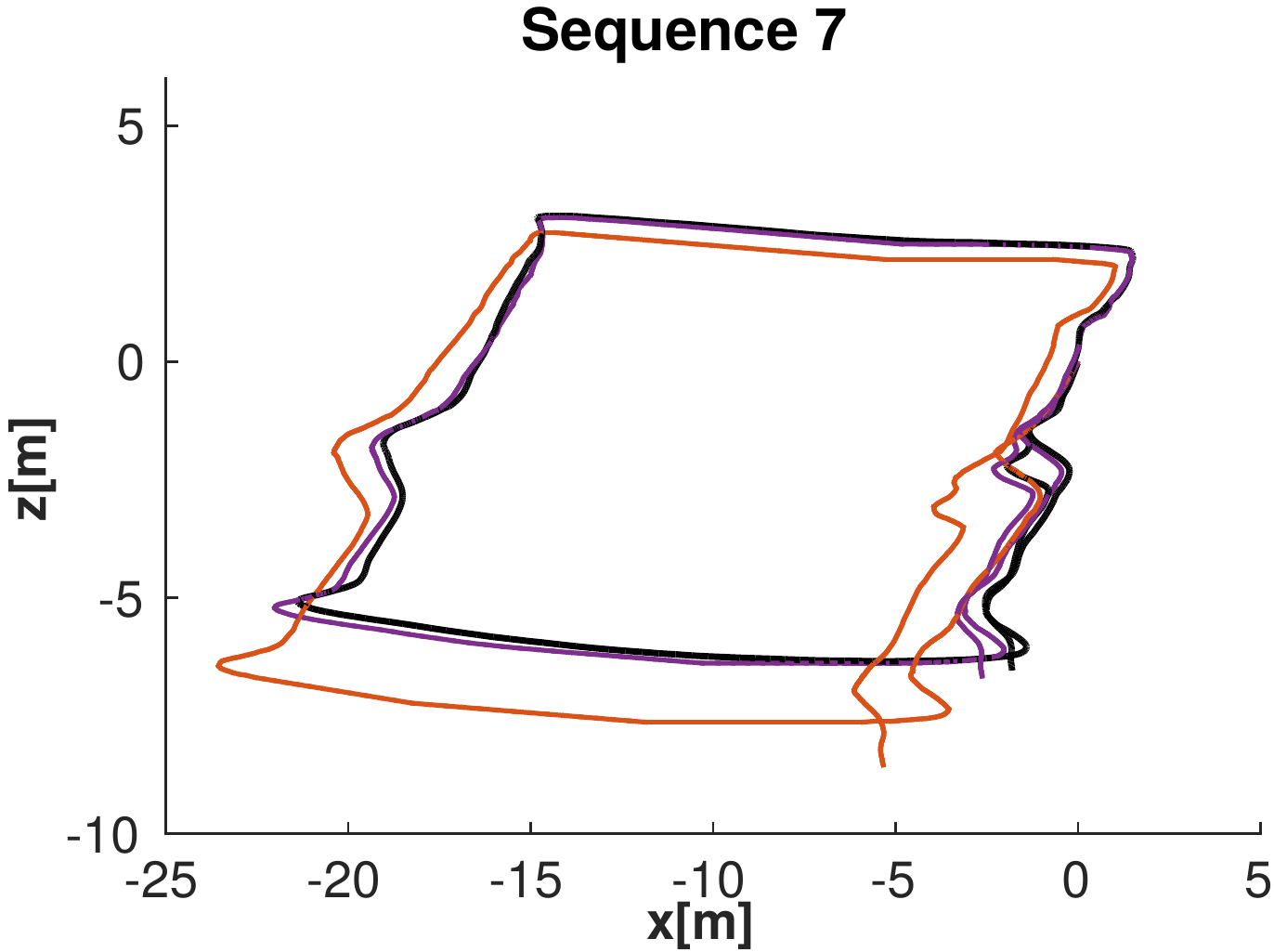}
\end{minipage}
\hspace{0.2cm}
\begin{minipage}[b]{0.25\textwidth}
	\centering
 	\includegraphics[width=\textwidth]{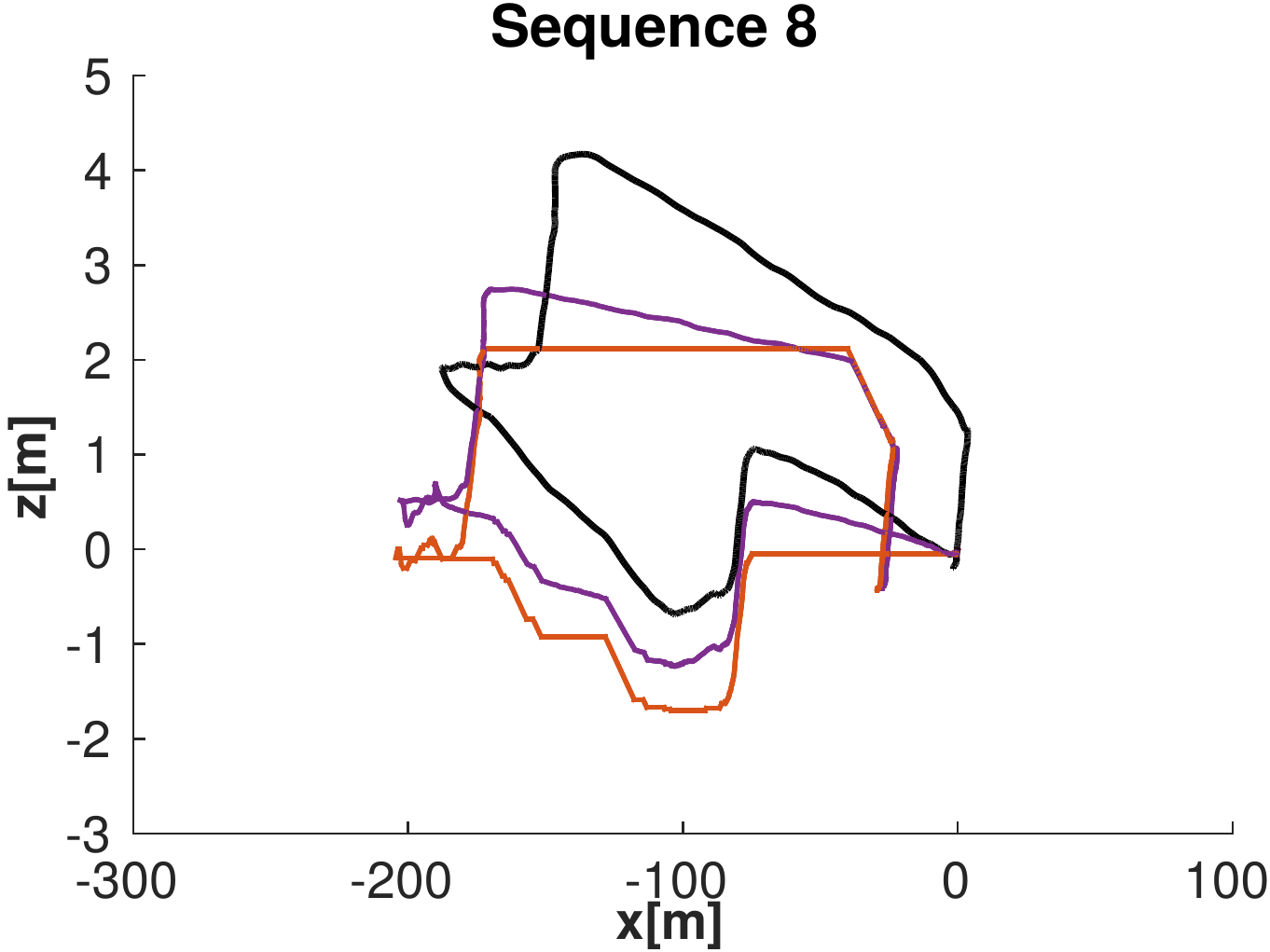}
\end{minipage}
\hspace{0.2cm}
\begin{minipage}[b]{0.25\textwidth}
	\centering
 	\includegraphics[width=\textwidth]{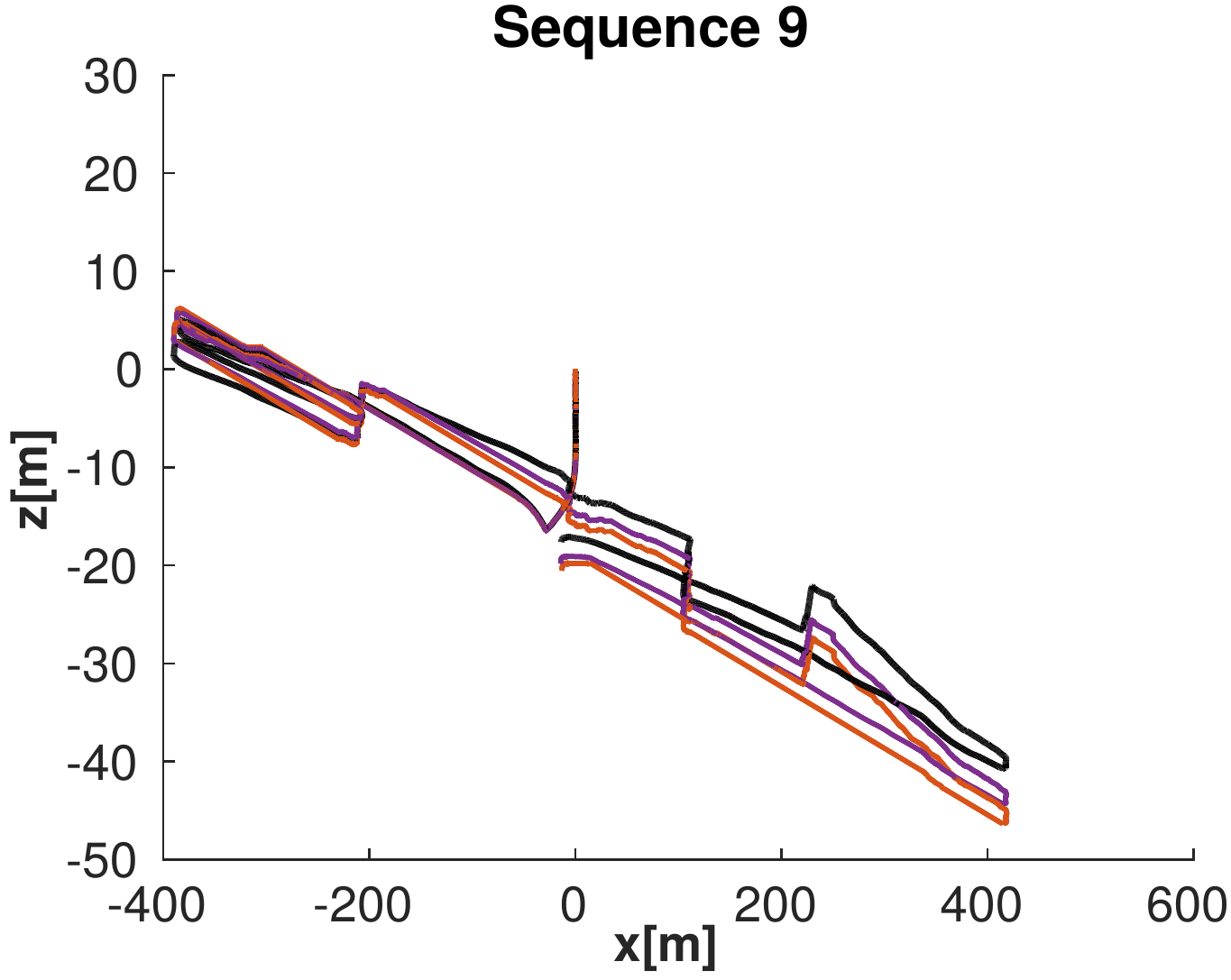}
\end{minipage}

\vspace{0.05cm}

\begin{minipage}[b]{0.25\textwidth}
	\centering
 	\includegraphics[width=\textwidth]{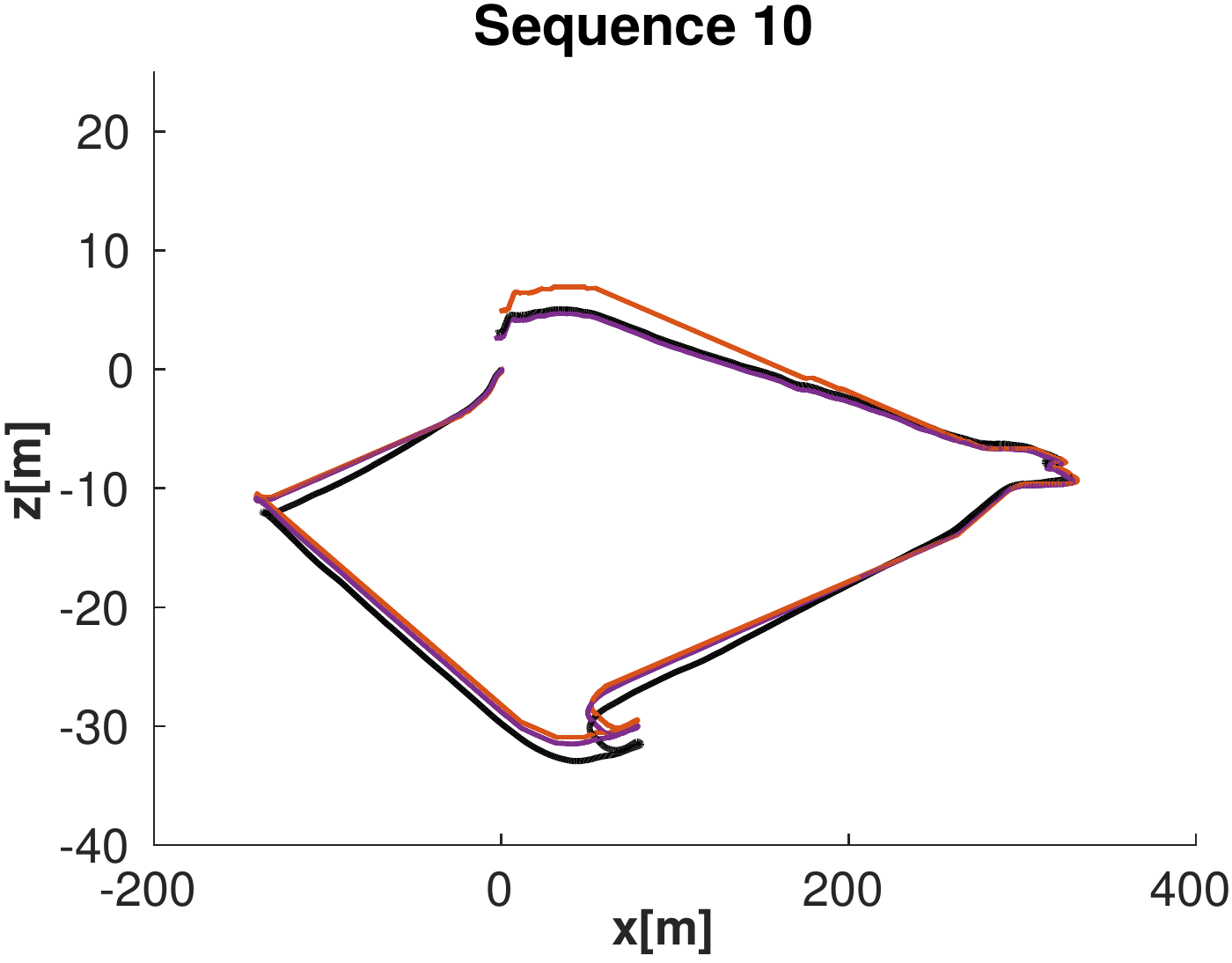}
\end{minipage}
\hspace{0.2cm}
\begin{minipage}[b]{0.25\textwidth}
	\centering
 	\includegraphics[width=\textwidth]{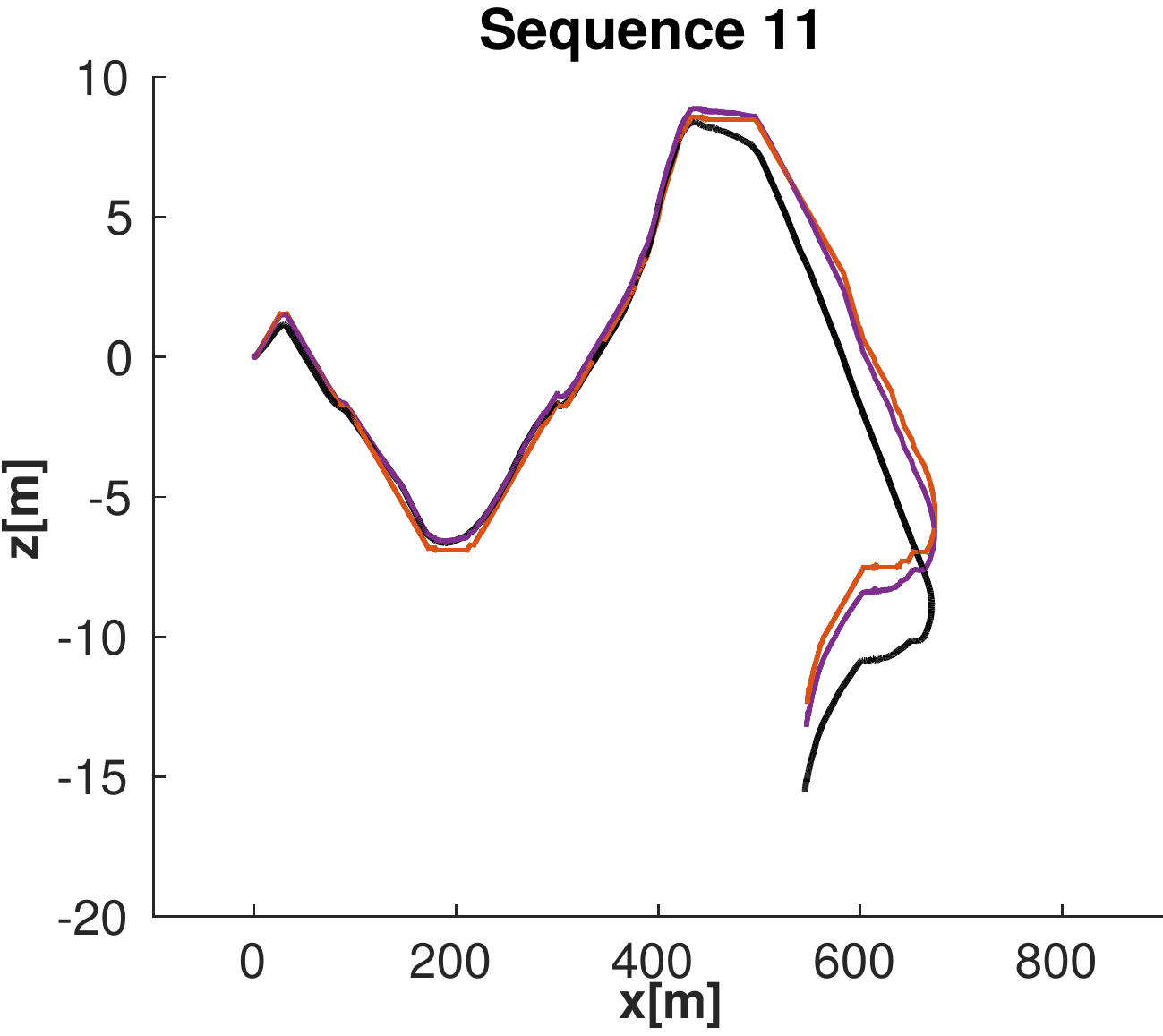}
\end{minipage}
\hspace{0.2cm}
\begin{minipage}[b]{0.15\textwidth}
	\centering
	\includegraphics[width=\textwidth]{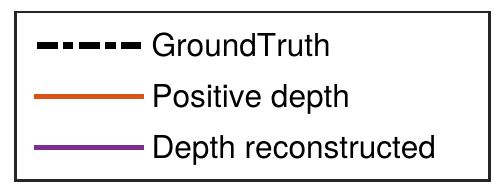}
\end{minipage}
\end{center}
\vspace*{-4mm}
\caption{Estimated paths projected onto the X-Z plane computed using our proposed method  for the 11 sequences of the Kitti dataset.  Results for sequence 5 and 8 show more distortion but, in general, the trajectories are successfully recovered, and the method that normalizes the estimation through the reconstruction of the 3D geometry of the scene achieves the best results.}
\label{fig:nflow_kittipath}
\end{figure}

Fig.~\ref{fig:nflow_kittipath} shows for the Kitti benchmark the estimated paths projected on the X-Z plane. For the scale, we used  the ground-truth translation speed. For each case, we show the trajectories for all the methods and the ground-truth. Most estimated trajectories are very similar to the ground-truth, and particularly after 3D reconstruction there is very high accuracy. However, for sequence 5 only the refined solution reaches a successful trajectory. This could be due to the small translational motion with respect to the rotational motion. Note the span of the X axis of less than 1 m compared to the hundreds of meters in all other sequences. Moreover, for sequence 8, all estimations deviate in the beginning, and the accumulated error prevents a recovery of the accurate ground-truth position.


\subsection{Comparison with direct methods}
\label{subsec:5.2}
\begin{table}[]
\centering
\caption{Translation AAE, Rotation AAE and EPE (${^\circ}$/frame)}
\label{tab:methods_normalflow}
   \scalebox{1.0}{
   		\resizebox{1\textwidth}{2.8cm}{
			\begin{tabular}{l||r|rr||r|rr||r|r}
			\hline
			 & \multicolumn{3}{c}{Yosemite} & \multicolumn{3}{c}{Fountain} & \multicolumn{2}{c}{Kitti 00 -10} \\
			 & Trans. AAE & Rot. AAE & Rot. EPE & Trans. AAE & Rot. AAE & Rot. EPE & Trans. AAE & Rot. EPE \\ \hline

			Raudies, 2014\cite{raudies_efficient_2009}
			using Brox\cite{brox_large_2011}& 			 2.4453 & 20.9280 & 0.0338 & 		1.6380 & 1.4605 & 0.0102 & 		 14.7759 & 0.4300\\
			Bruss, 1983\cite{bruss_passive_1983}
			using Brox\cite{brox_large_2011}& 			 1.5285 & 28.5546 & 0.0126 &  		1.1601 & 3.0578 & \textbf{0.0028} & 		13.8722 & 0.3690\\
			Kanatani, 1993\cite{kanatani_3dinterpretation_1993}
			using Brox\cite{brox_large_2011}& 			 2.1252 & 13.8169 & 0.0182 & 		1.4726 & 1.6544 & 0.0071 & 		 14.5035 & 0.4781\\

			Raudies, 2014\cite{raudies_efficient_2009}
			using Sun\cite{sun_quantitative_2014}& 			2.2446 & 19.1784 & 0.0340 & 		 0.5120 & 0.5881 & \textbf{0.0026} & 		 14.0957 & 0.4504\\
			Bruss, 1983\cite{bruss_passive_1983}
			using Sun\cite{sun_quantitative_2014}& 			1.3245 & 26.7049 & 0.0125 &  		 1.0248 & 2.0480 & 0.0034 & 		13.4256 & 0.3689\\
			Kanatani, 1993\cite{kanatani_3dinterpretation_1993}
			using Sun\cite{sun_quantitative_2014}& 			1.9742 & 10.8016 & 0.0122 & 		 \textbf{0.4012} & \textbf{0.5607} & 0.0035 & 		 13.8342 & 0.4928\\

			Raudies, 2014\cite{raudies_efficient_2009}
			using Vogel\cite{vogel_evaluation_2013}& 			0.8085 & 21.5764 & 0.0109 & 		 0.7657 & 0.7893 & 0.0059 & 		13.6363 & 0.4338\\
			Bruss, 1983\cite{bruss_passive_1983}
			using Vogel\cite{vogel_evaluation_2013}& 			1.2289 & 26.4552 & 0.0122 &  		 1.5791 & 3.4248 & 0.0030 & 		12.5186 & 0.1214\\
			Kanatani, 1993\cite{kanatani_3dinterpretation_1993}
			using Vogel\cite{vogel_evaluation_2013}& 			0.7866 & 19.7755 & 0.0089 & 		 0.6864 & 1.0900 & 0.0041 & 		13.4672 & 0.6690\\\hline

			Hui, 2013\cite{hui_structure_2013} & 		NA & NA & NA &		 			 2.497 & 3.907 & NA &  				 NA & NA \\
			Yuan, 2013\cite{yuan_direct_2013} & 		NA & NA & NA & 					 1.251 & NA & 0.0699 &  				 NA & NA \\
			Hui, 2015\cite{hui_determining_2015} & 		1.619 & NA & 0.0282 & 			 2.371 & NA & 0.0220 &  				 NA & NA \\
			Yuan, 2015\cite{yuan_camera_2015} & 		0.8803 & NA & 0.0685 & 			 1.1866 & NA & 0.050 &  				 NA & NA \\
			Ji, 2006\cite{ji_3dshape_2006} & 	0.9589 & 5.4261 & \textbf{0.0054} &		1.4043 & 1.9327 & \textbf{0.0022} & 			2.4923 & 0.0803 \\

			\textbf{Our approach: positive depth} & 	0.8436 & \textbf{4.9055} & 0.3799 & 			 1.2054 & 1.3528 & 0.2138 &  				 1.8225 & \textbf{0.0613} \\
			\textbf{Our approach: depth reconstructed} & 	\textbf{0.3640} & 5.7749 & 0.3789 & 			 \textbf{0.4074} & \textbf{0.5615} & 0.2139 &  				 \textbf{1.5340} & 0.3545 \\\hline
			\end{tabular}
		}
	}
\end{table}
Table \ref{tab:methods_normalflow} summarizes the comparison to several works in the literature using both normal flow and optical flow.
The first rows show the error for three optical flow based methods: Raudies~\cite{raudies_review_2012}, Bruss~\cite{bruss_passive_1983}, and Kanatani~\cite{kanatani_3dinterpretation_1993}. The optical flow is computed using three different methods: Brox~\cite{brox_large_2011}, Sun~\cite{sun_quantitative_2014}, and Vogel~\cite{vogel_evaluation_2013}.

The optical flow method of a) Sun \cite{sun_quantitative_2014}, which ranked top 1 in 2014, is a variation of Horn \& Schunck \cite{horn_determining_1981}. It uses a non-local smoothness regularization term based on median filtering, includes boundary and occlusion prediction to preserve motion, and uses an asymmetric hierarchical pyramid strategy to improve large motion estimations. b) The method of Vogel \cite{vogel_evaluation_2013} is a variational method that uses Total Generalized Variation (TGV) regularization with a data term based on Census Transform with convex optimization (CSAD). c) The Brox \cite{brox_large_2011} method targets specifically large displacements and is well suited for the Kitti dataset, where the inter-frame displacements reach up to 50 pixels.

The three selected optical-flow based methods for 3D motion estimation have been introduced in Section \S\ref{sec:1}. These three methods showed the best performance and consistency in our exhaustive comparison in the lab.
They have been included to facilitate a comparison with direct approaches and seeking completeness for future comparisons.

The next rows show methods that use normal flow for estimating ego-motion. Although, other methods exist in the literature (\cite{fermuller_passive_1995,brodsky_structure_2000,silva_robust_1997}), we only included methods which have been evaluated on modern datasets. Hui \etal \cite{hui_structure_2013} compute 3D motion from monocular normal flows and then estimate depth from a stereo system without explicitly computing correspondences. They use two constraints: the AFD, which is a relaxation of the constraint in Eq.~(\ref{eq:voting_robust}) where a voting mechanism punishes the estimates that result in negative depth (different signs for the translational and rotational components).
 To make the constraint more robust, several re-projections from one camera on the other are also considered. The second constraint, called AFM, affects the motion magnitude, and is similar to the constraint in Eq.~(\ref{eq:constraint_normal_perpendicular_transflow}), that uses normal flows orthogonal to the translational component.
 In \cite{hui_determining_2015} the same authors propose a new solution based on the AFD constraint.

The method of Yuan \etal \cite{yuan_direct_2013} first estimates the 3D motion using the constraint in Eq.~(\ref{eq:constraint_normal_perpendicular_transflow}), selecting the normal flow vectors orthogonal to the translational motion component. Then, they solve for the rotation with a more robust strategy, removing only solutions that do not lie in the  half-plane consistent with the normal flow estimates. In Yuan \etal \cite{yuan_camera_2015} authors present an updated version that relies on clustering for selecting the flows that satisfy the earlier described constraint and adding a step using RANSAC to refine the final estimations.
We also have re-implemented the patch-based method for a single normal flow field described in \cite{ji_3dshape_2006}. This method, partitions the image into regular patches, and models the scene of each patch with a plane. We note that this planar constraint increases the number of parameters to be estimated.

 Referring to the Table, our methods outperforms all normal flow based methods in all cases (cases for which results were not available are marked as NA). Our method also outperforms all optical flow based methods, but the method of Kanatani \cite{kanatani_3dinterpretation_1993} using the optical flow from Sun \cite{sun_quantitative_2014} on the the Fountain sequence. The results of our method may be further improved with an additional refinement step of the rotation after refining the translation, but at the cost of increased computation. We note the high accuracy of the method in \cite{ji_3dshape_2006}, specifically for the rotational velocity. Let us emphasize the high performance of our method on the Kitti dataset, which is a real-world driving scenario in contrast to all other synthetic datasets previously used in the evaluation of direct 3D motion estimation methods.

\subsection{3D reconstruction from motion}
\label{subsec:5.3}
\begin{figure*}[tb]
\begin{center}	
	\vspace{-1.5cm}
	\begin{minipage}[b]{0.20\textwidth}
		\centering
 		\includegraphics[width=\textwidth, height=2.1cm]{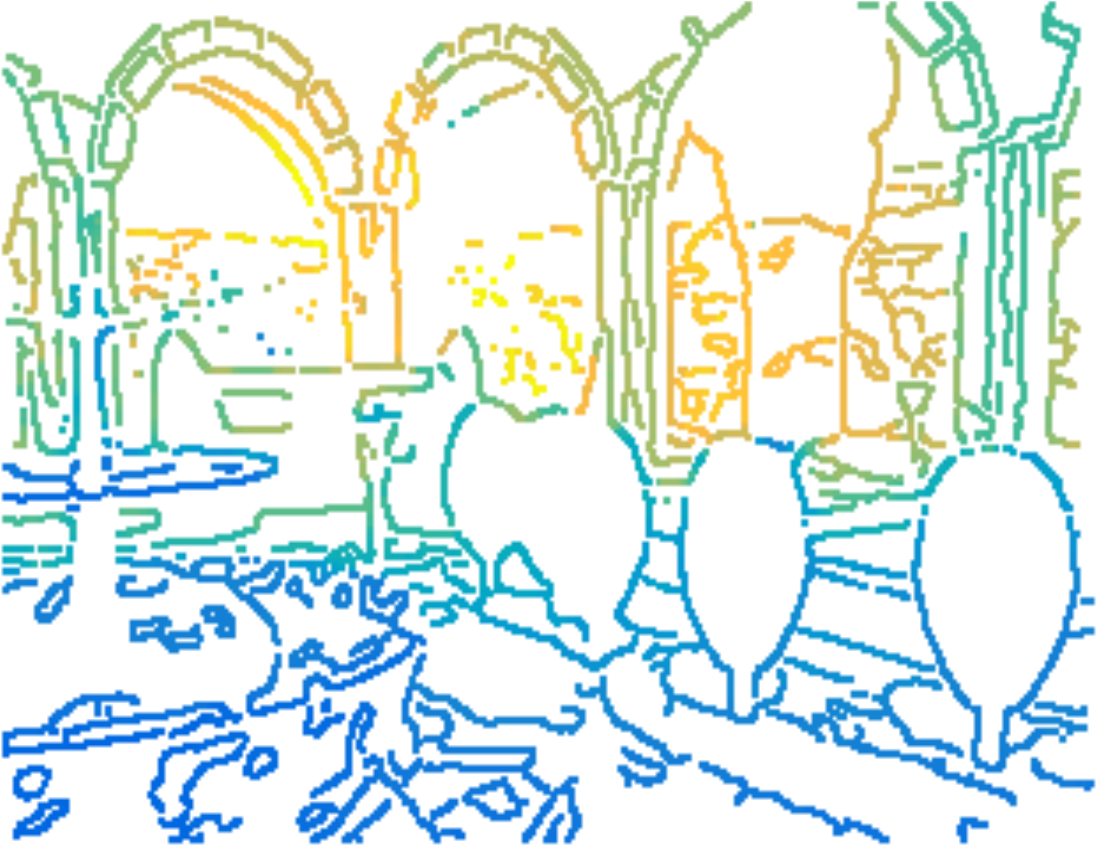}
	\end{minipage}
	\hspace{0.0025cm}
	\begin{minipage}[b]{0.20\textwidth}
		\centering
	 	\includegraphics[width=\textwidth, height=2.1cm]{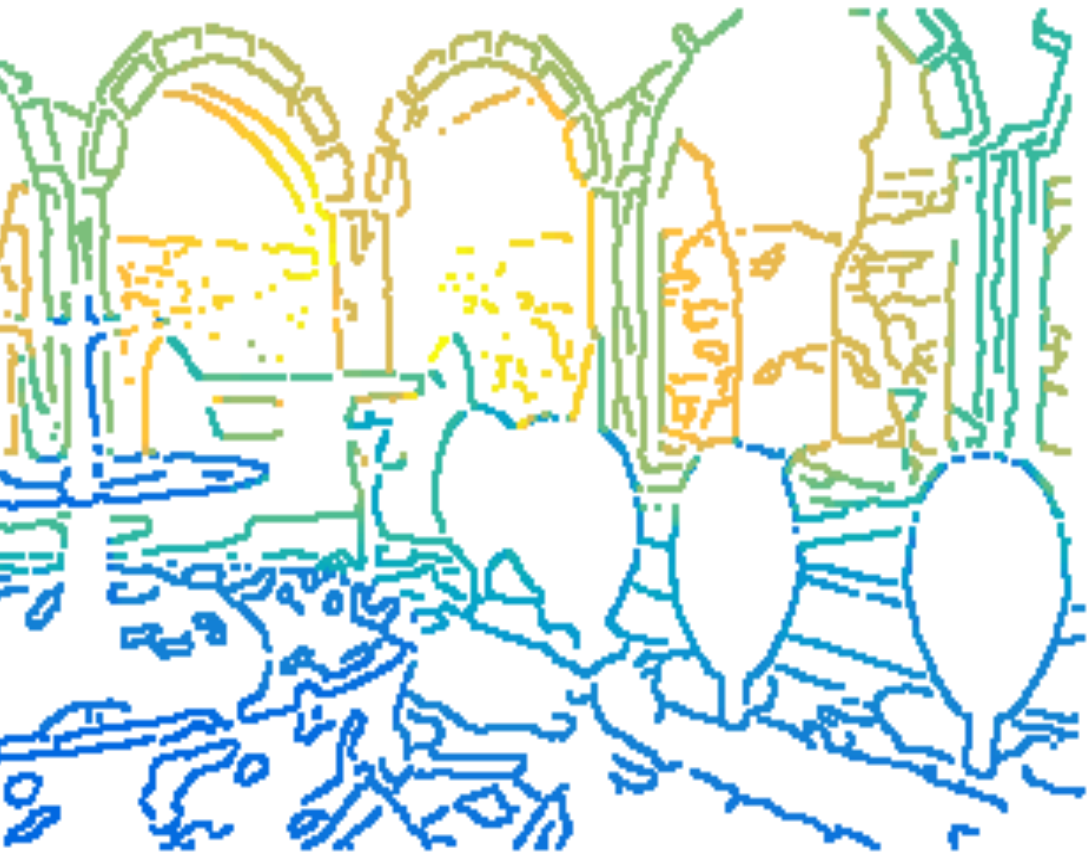}
	\end{minipage}
	\hspace{0.005cm}
	\begin{minipage}[b]{0.26\textwidth}
		\centering
 		\includegraphics[width=\textwidth, height=2.1cm]{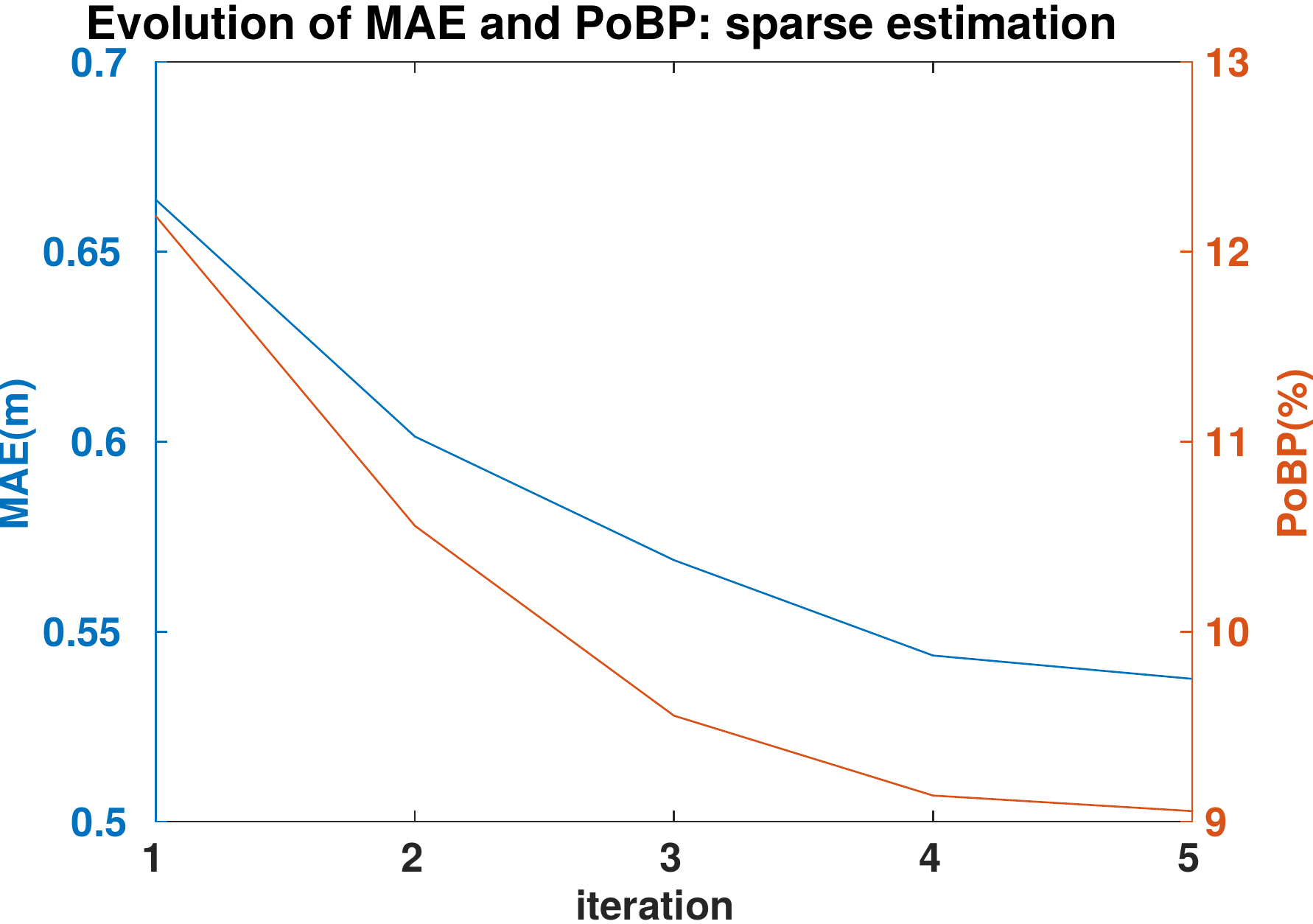}
	\end{minipage}
	\hspace{0.0025cm}	
	\begin{minipage}[b]{0.26\textwidth}
		\centering
 		\includegraphics[width=\textwidth, height=2.1cm]{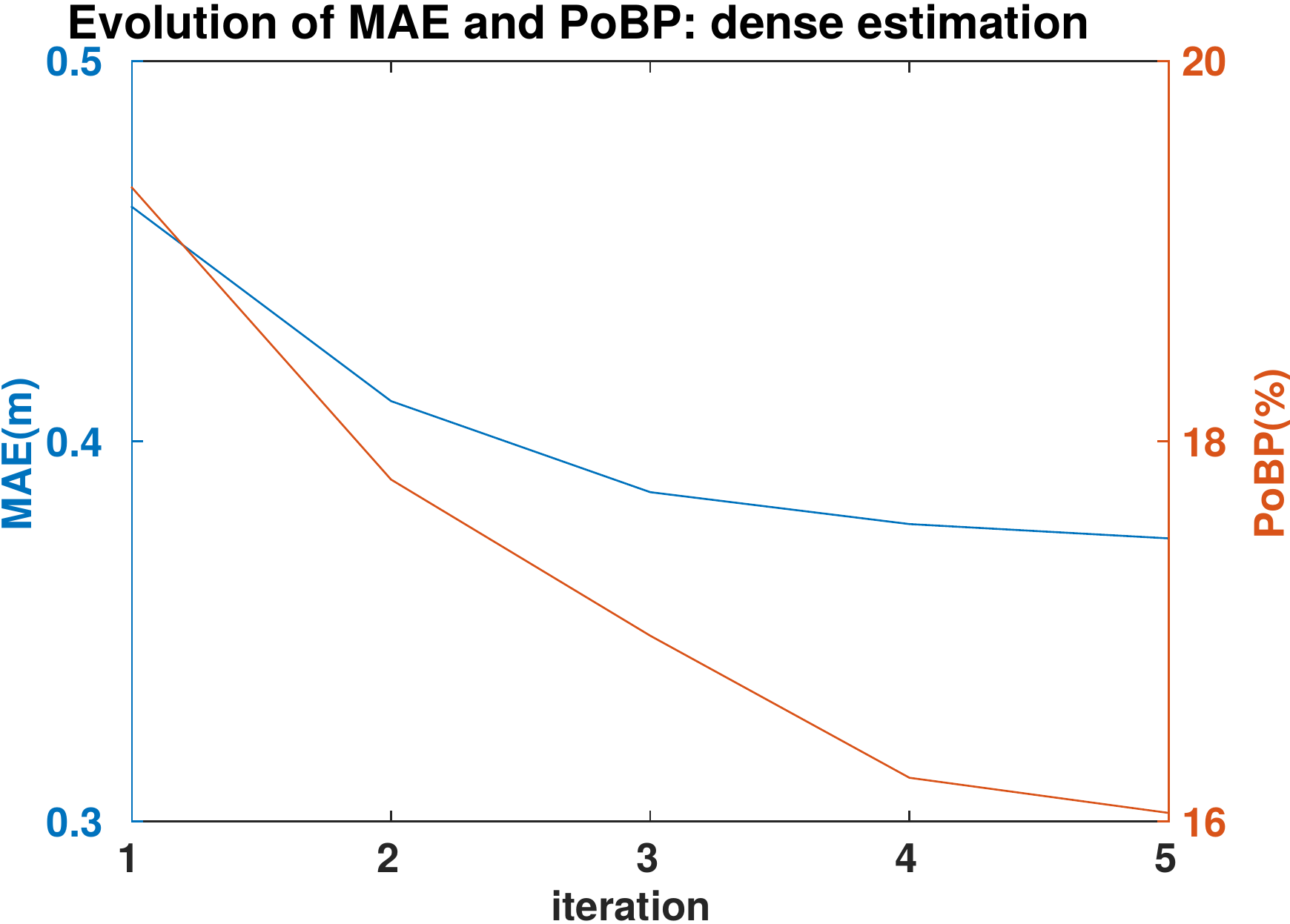}
	\end{minipage}
	
%
%
	
	\begin{minipage}[b]{0.42\textwidth}
		\begin{minipage}[b]{\textwidth}		
		\centering
 		\includegraphics[width=\textwidth, height=1.2cm]{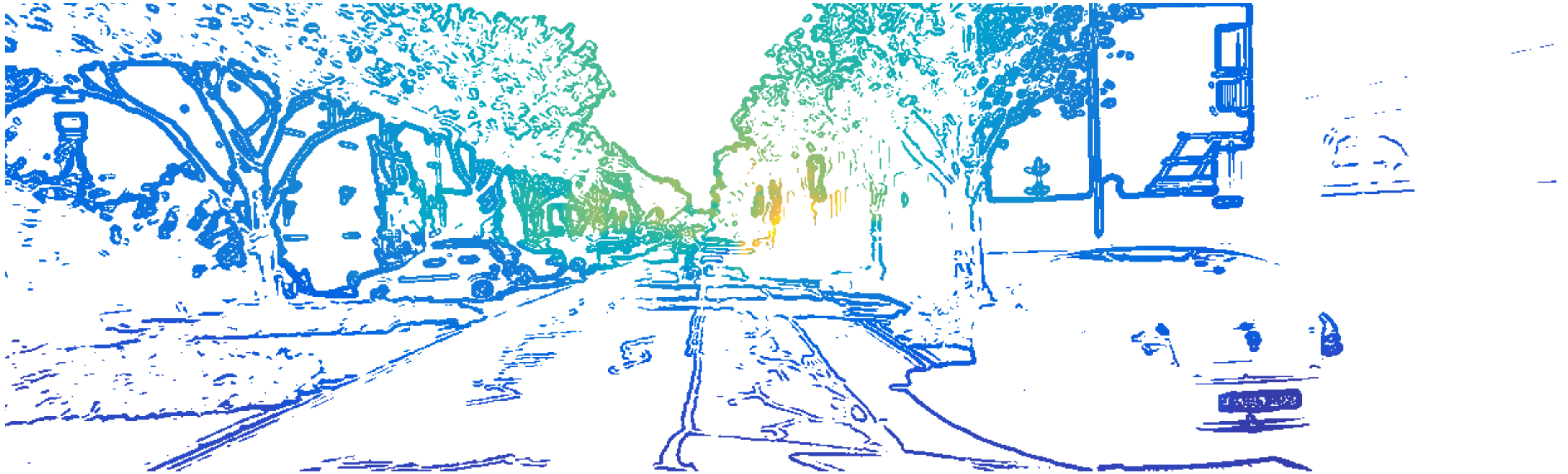}
 		\end{minipage}
 		\begin{minipage}[b]{\textwidth}		
		\centering
 		\includegraphics[width=\textwidth, height=1.2cm]{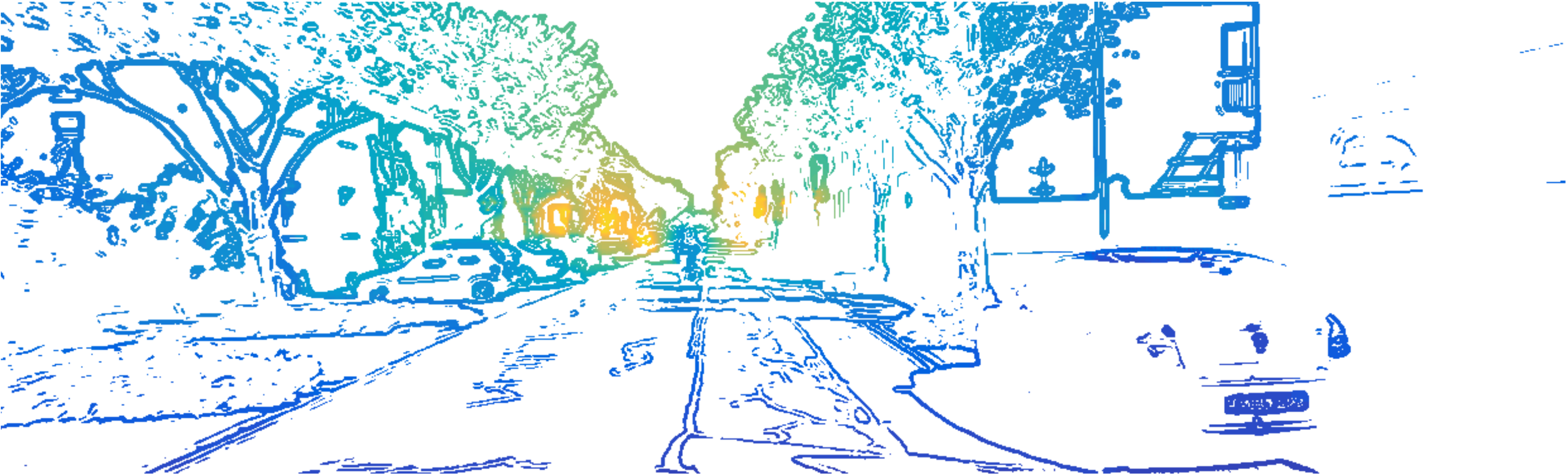}
 		\end{minipage}
	\end{minipage}
	\hspace{0.005cm}
	\begin{minipage}[b]{0.26\textwidth}
		\centering
 		\includegraphics[width=\textwidth, height=2.4cm]{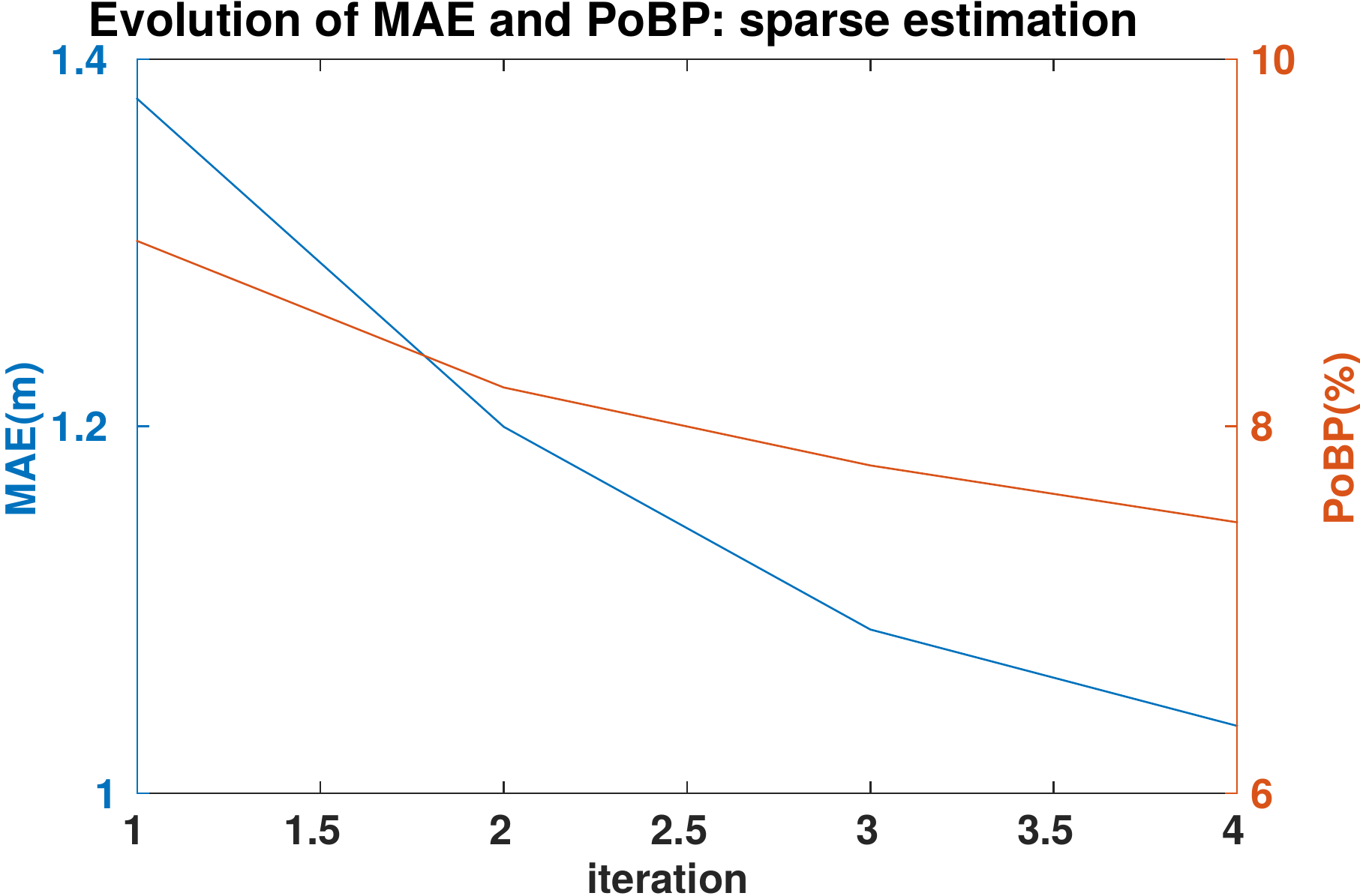}
	\end{minipage}
	\hspace{0.0025cm}	
	\begin{minipage}[b]{0.26\textwidth}
		\centering
 		\includegraphics[width=\textwidth, height=2.4cm]{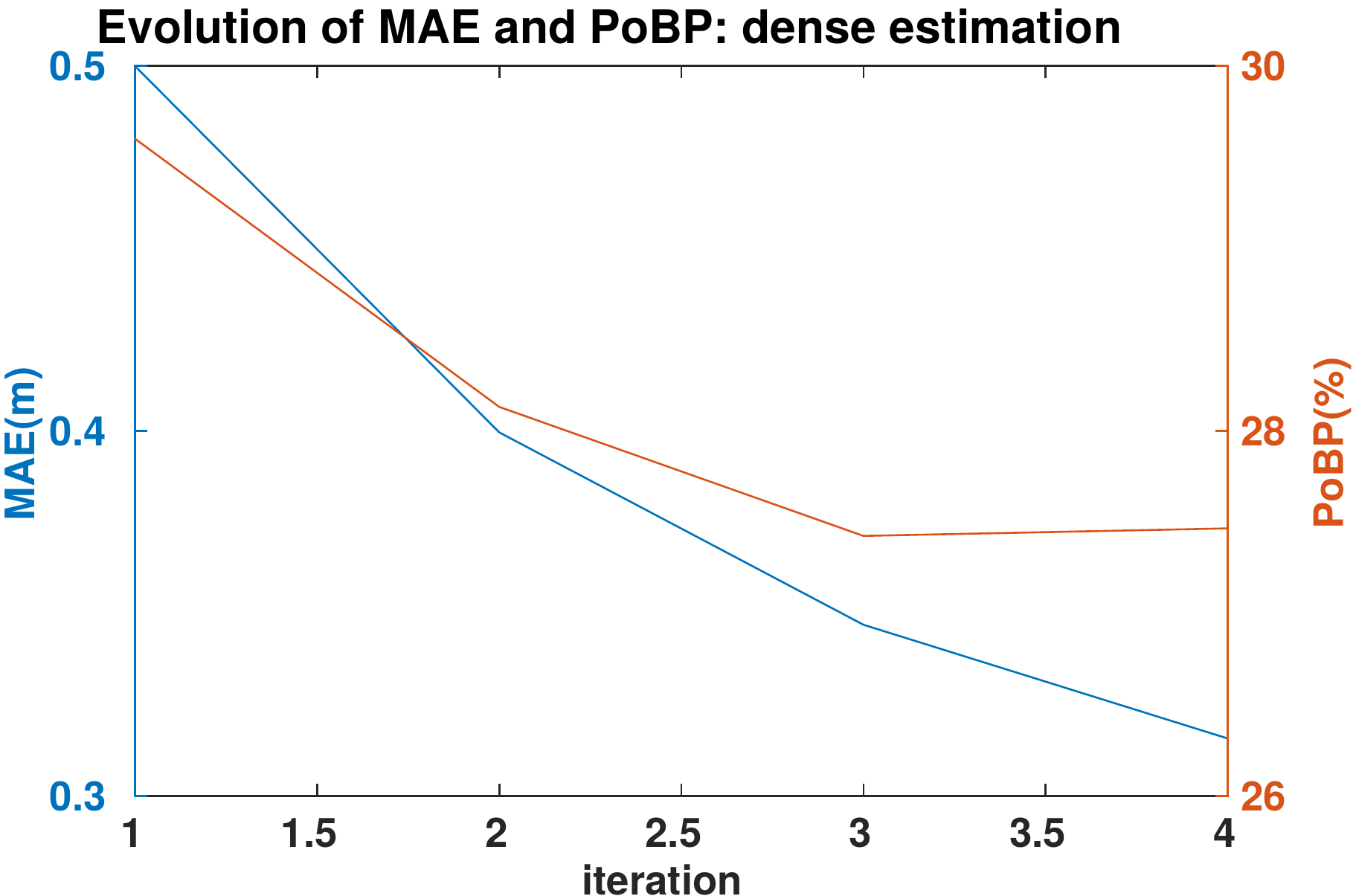}
	\end{minipage}
	
	\vspace{0.025cm}
	
	\begin{minipage}[b]{0.42\textwidth}
		\begin{minipage}[b]{\textwidth}		
		\centering
 		\includegraphics[width=\textwidth, height=1.2cm]{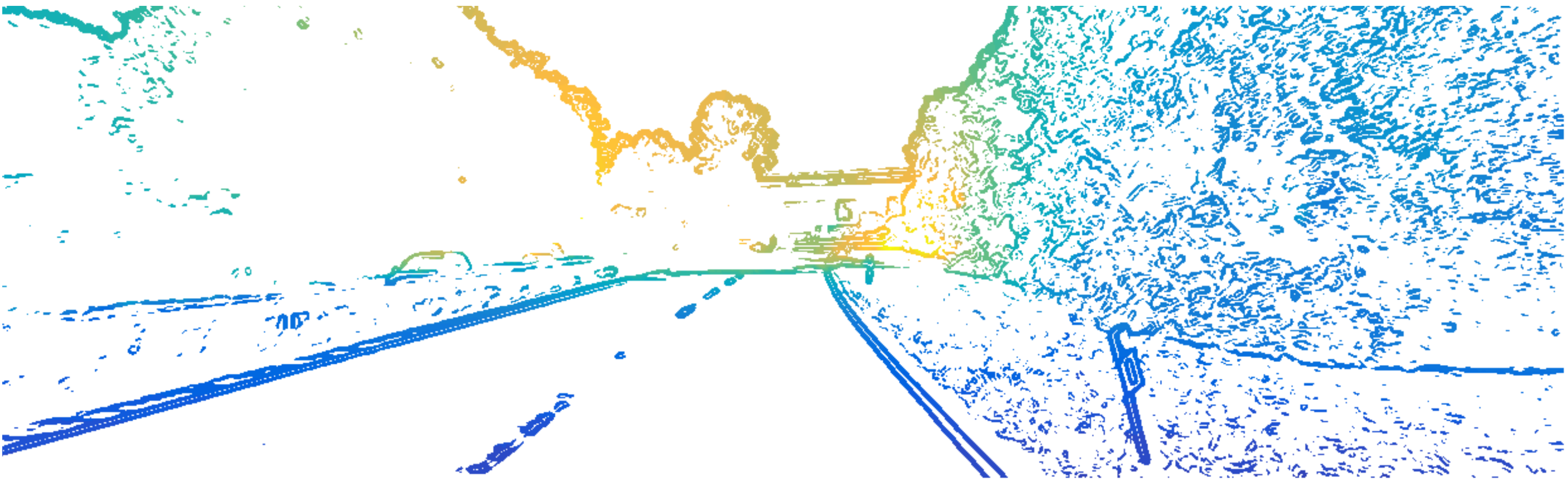}
 		\end{minipage}
 		\begin{minipage}[b]{\textwidth}		
		\centering
 		\includegraphics[width=\textwidth, height=1.2cm]{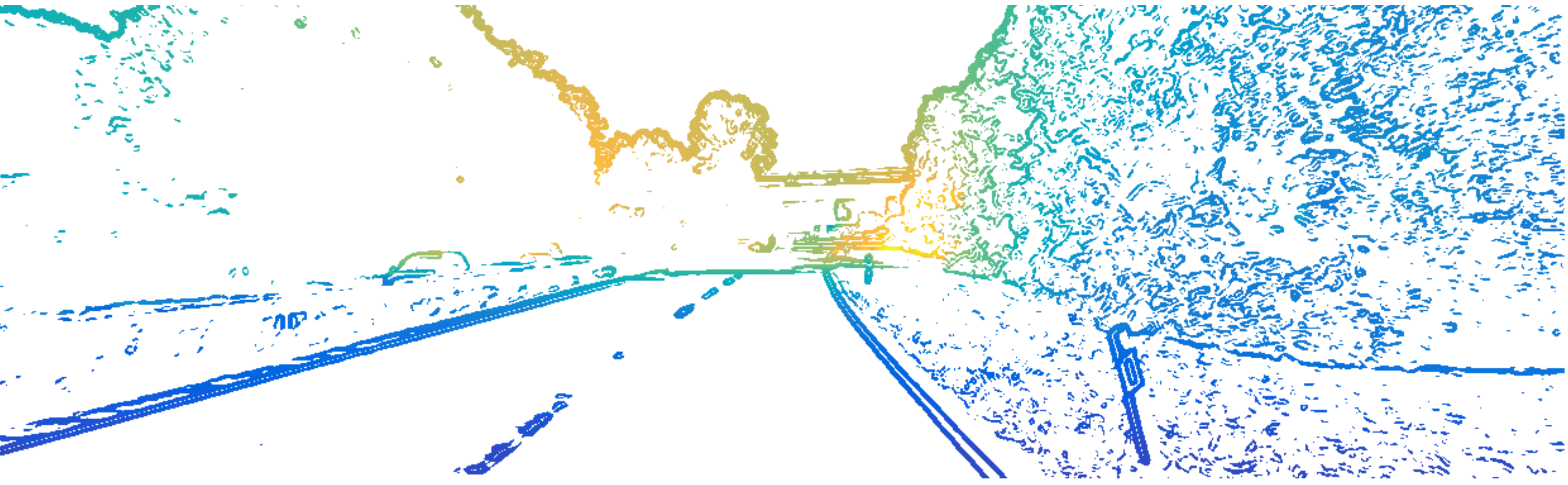}
 		\end{minipage}
	\end{minipage}
	\hspace{0.005cm}
	\begin{minipage}[b]{0.26\textwidth}
		\centering
 		\includegraphics[width=\textwidth, height=2.4cm]{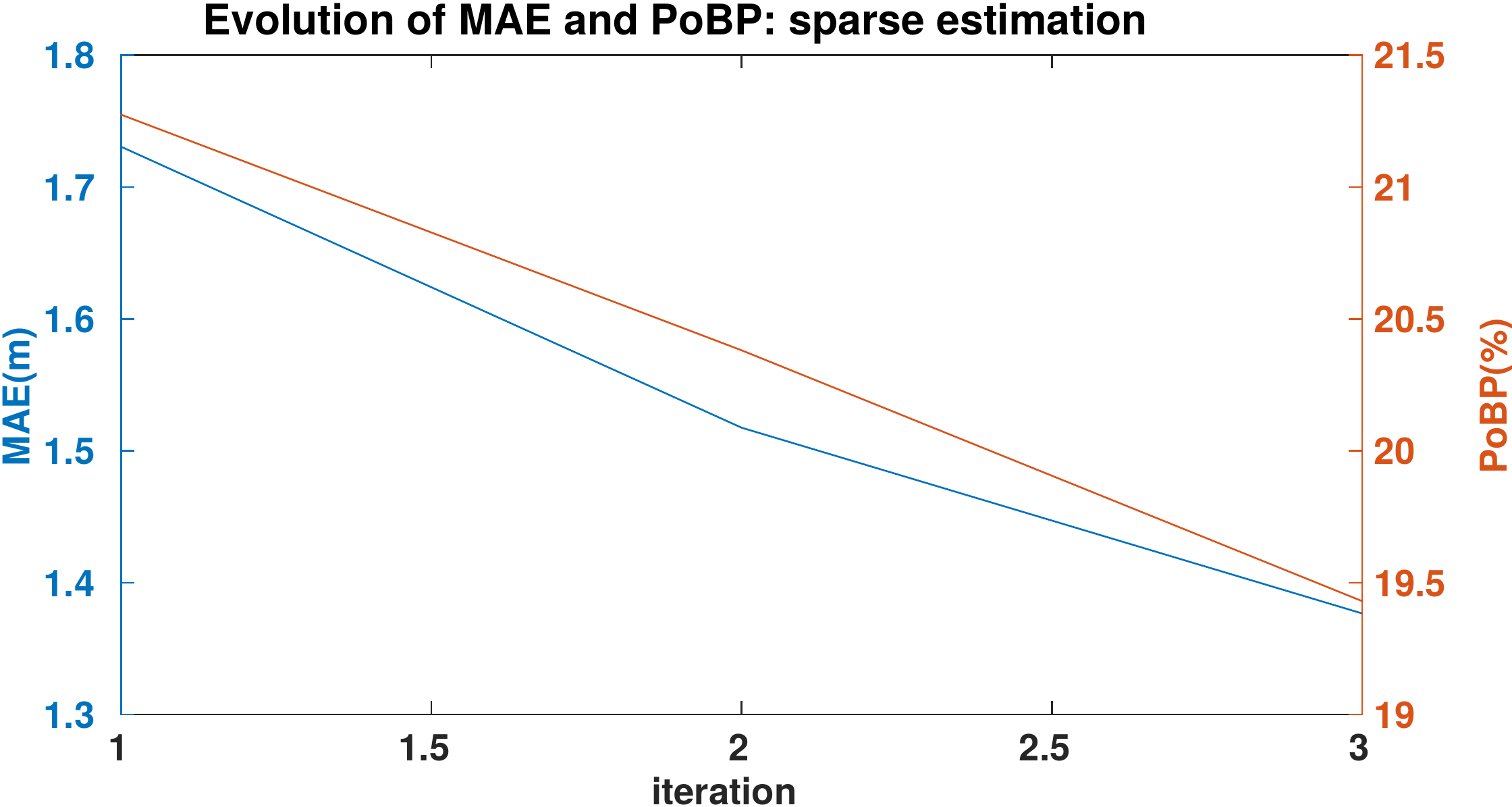}
	\end{minipage}
	\hspace{0.0025cm}	
	\begin{minipage}[b]{0.26\textwidth}
		\centering
 		\includegraphics[width=\textwidth, height=2.4cm]{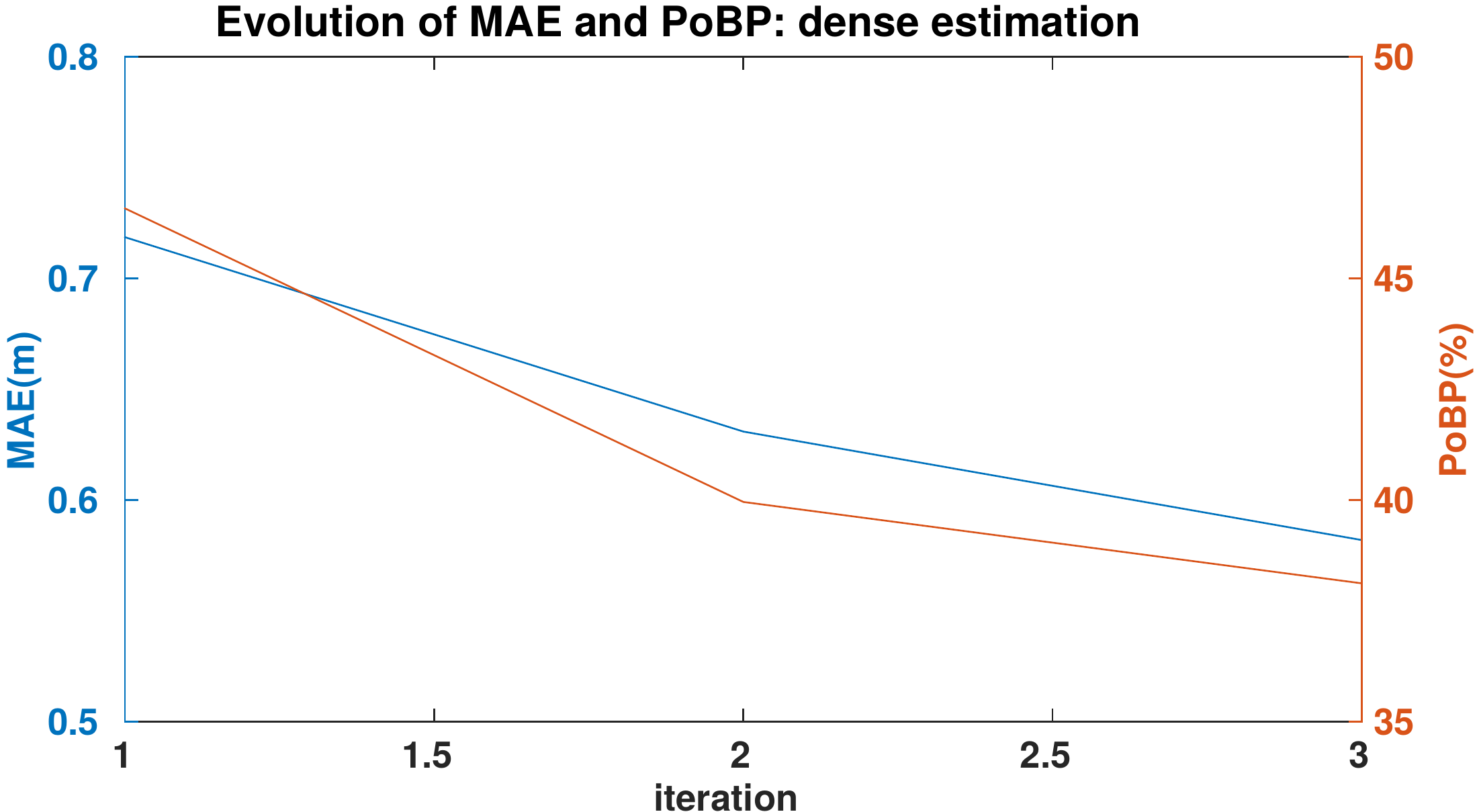}
	\end{minipage}
	
	\vspace{0.025cm}
	
	\begin{minipage}[b]{0.42\textwidth}
		\begin{minipage}[b]{\textwidth}		
		\centering
 		\includegraphics[width=\textwidth, height=1.2cm]{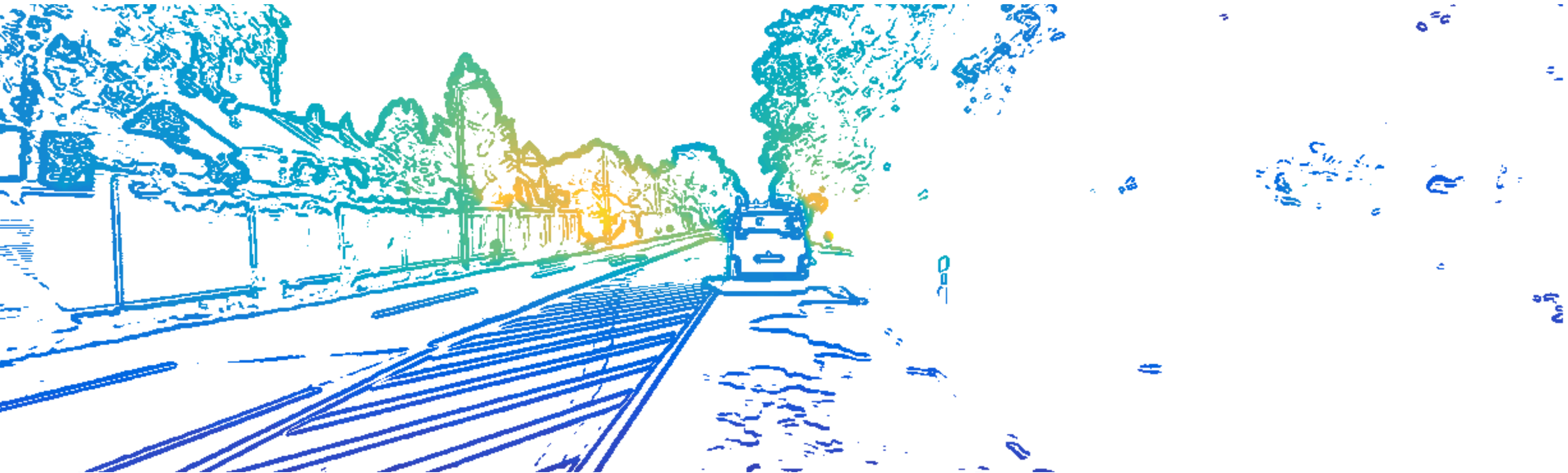}
 		\end{minipage}
 		\begin{minipage}[b]{\textwidth}		
		\centering
 		\includegraphics[width=\textwidth, height=1.2cm]{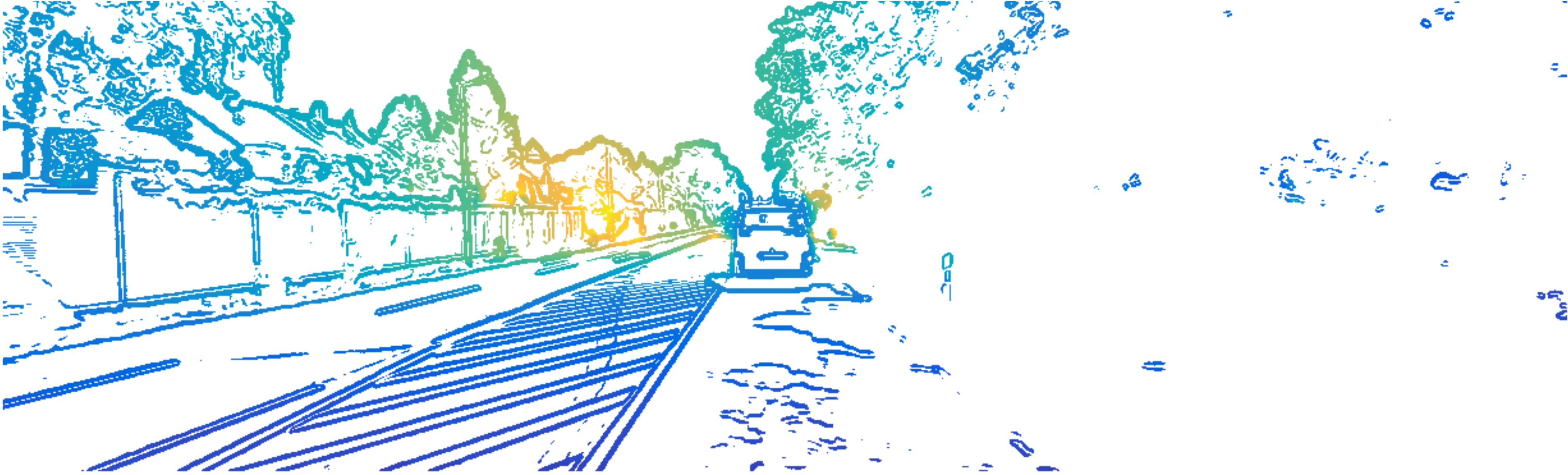}
 		\end{minipage}
	\end{minipage}
	\hspace{0.005cm}
	\begin{minipage}[b]{0.26\textwidth}
		\centering
 		\includegraphics[width=\textwidth, height=2.4cm]{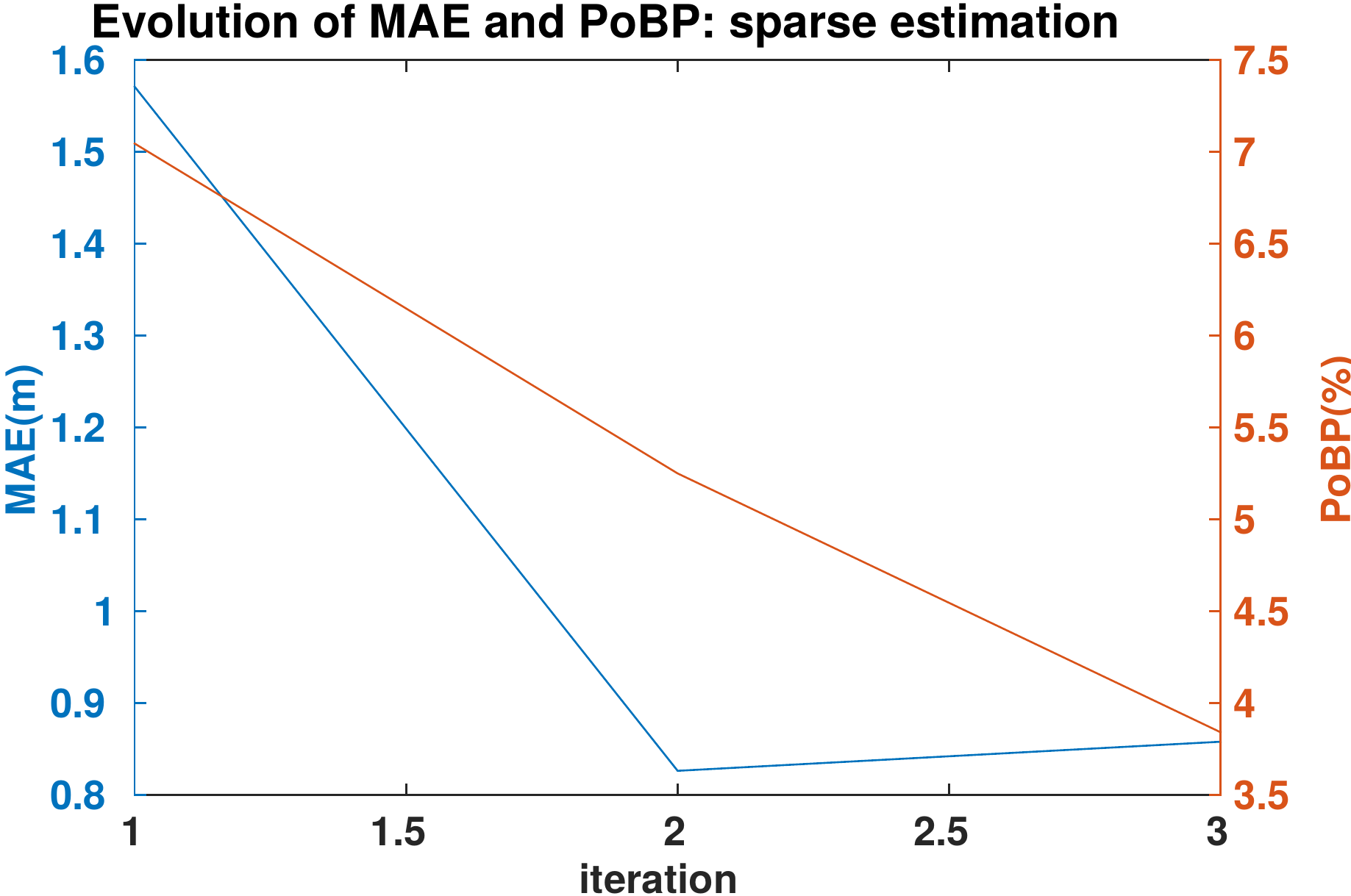}
	\end{minipage}
	\hspace{0.0025cm}	
	\begin{minipage}[b]{0.26\textwidth}
		\centering
 		\includegraphics[width=\textwidth, height=2.4cm]{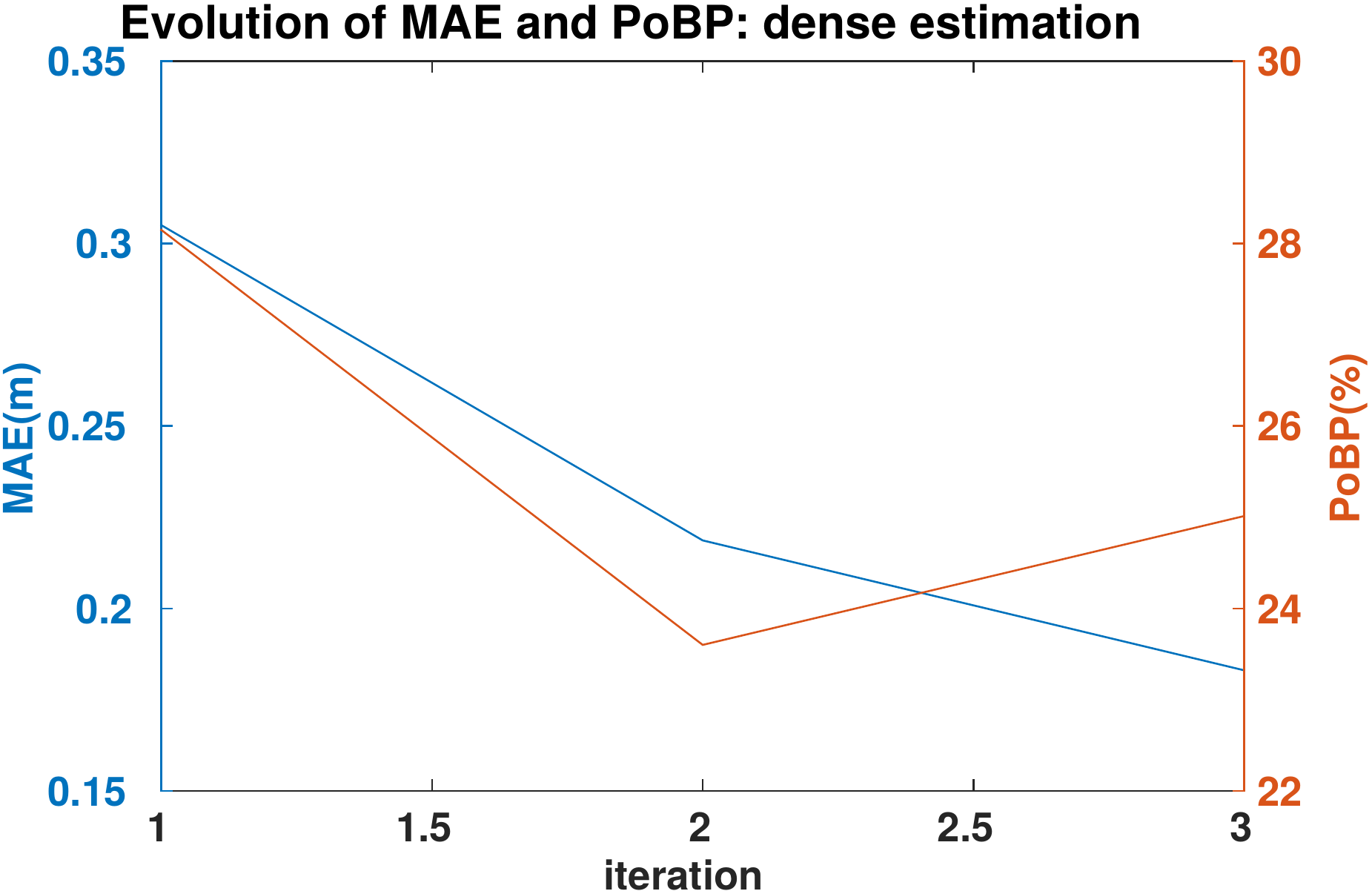}
	\end{minipage}
	
	\vspace{0.025cm}
	
	\begin{minipage}[b]{0.42\textwidth}
		\begin{minipage}[b]{\textwidth}		
		\centering
 		\includegraphics[width=\textwidth, height=1.2cm]{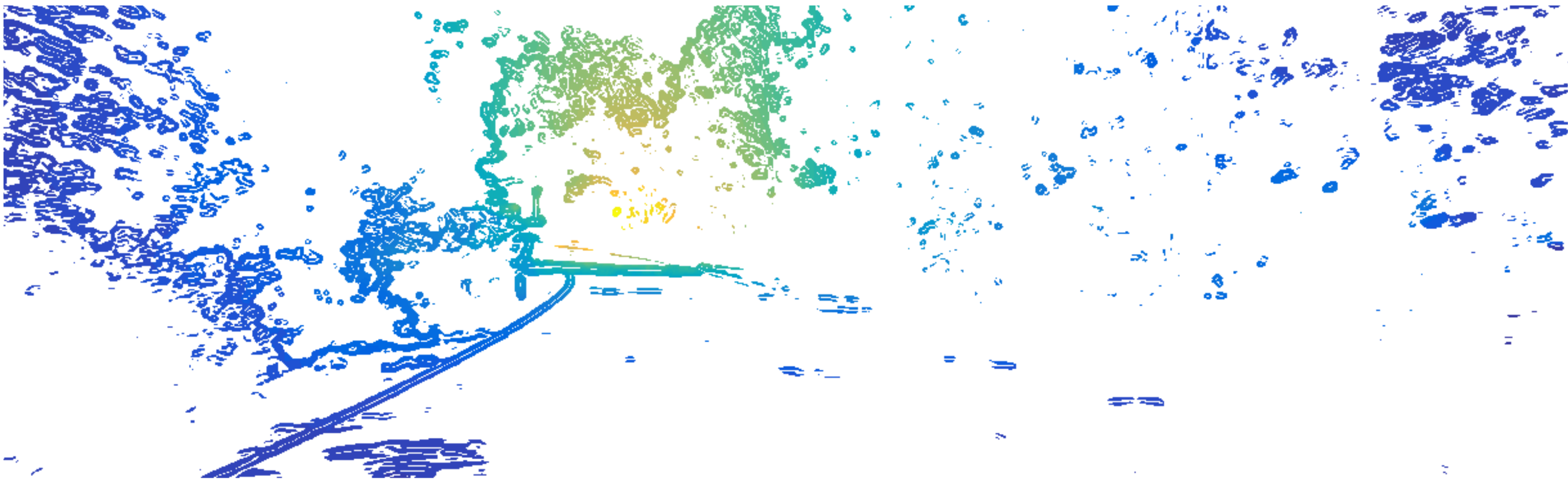}
 		\end{minipage}
 		\begin{minipage}[b]{\textwidth}		
		\centering
 		\includegraphics[width=\textwidth, height=1.2cm]{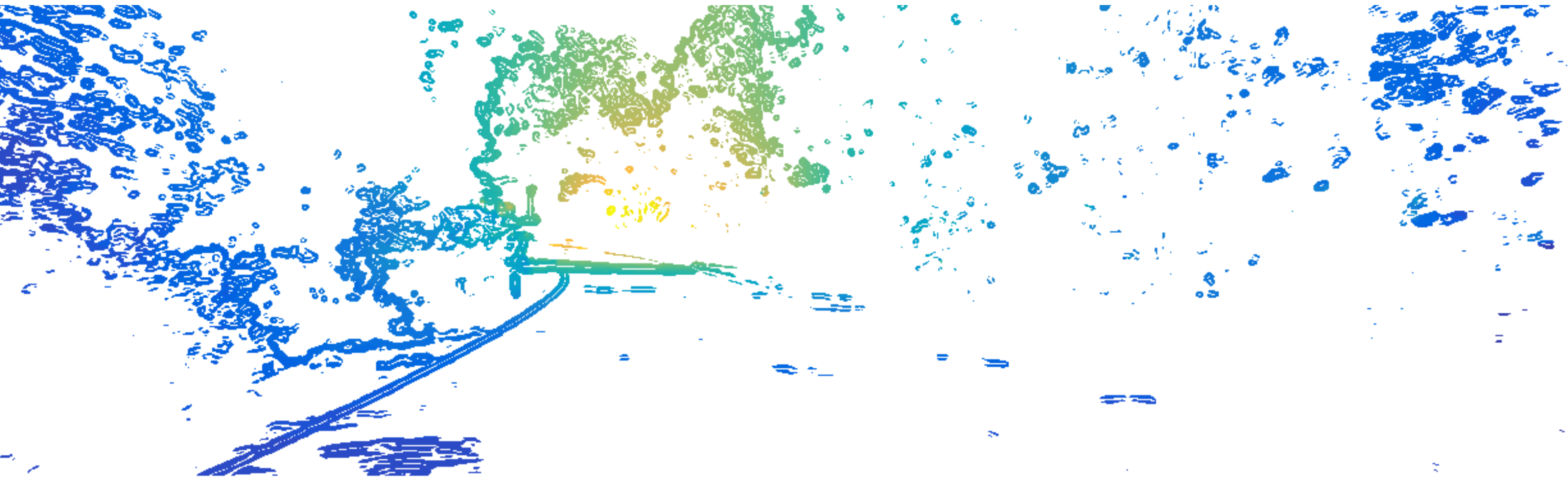}
 		\end{minipage}
	\end{minipage}
	\hspace{0.005cm}
	\begin{minipage}[b]{0.26\textwidth}
		\centering
 		\includegraphics[width=\textwidth, height=2.4cm]{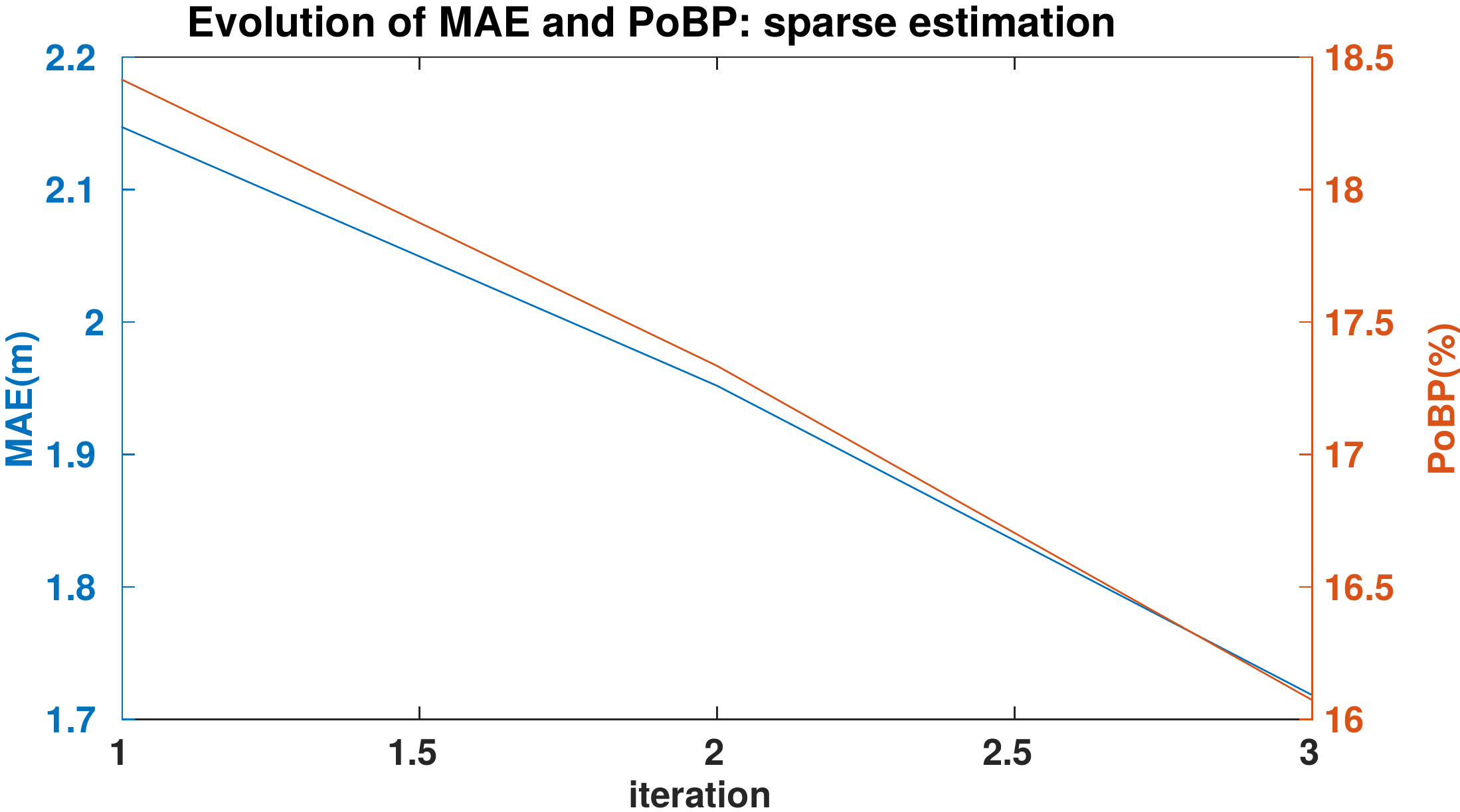}
	\end{minipage}
	\hspace{0.0025cm}	
	\begin{minipage}[b]{0.26\textwidth}
		\centering
 		\includegraphics[width=\textwidth, height=2.4cm]{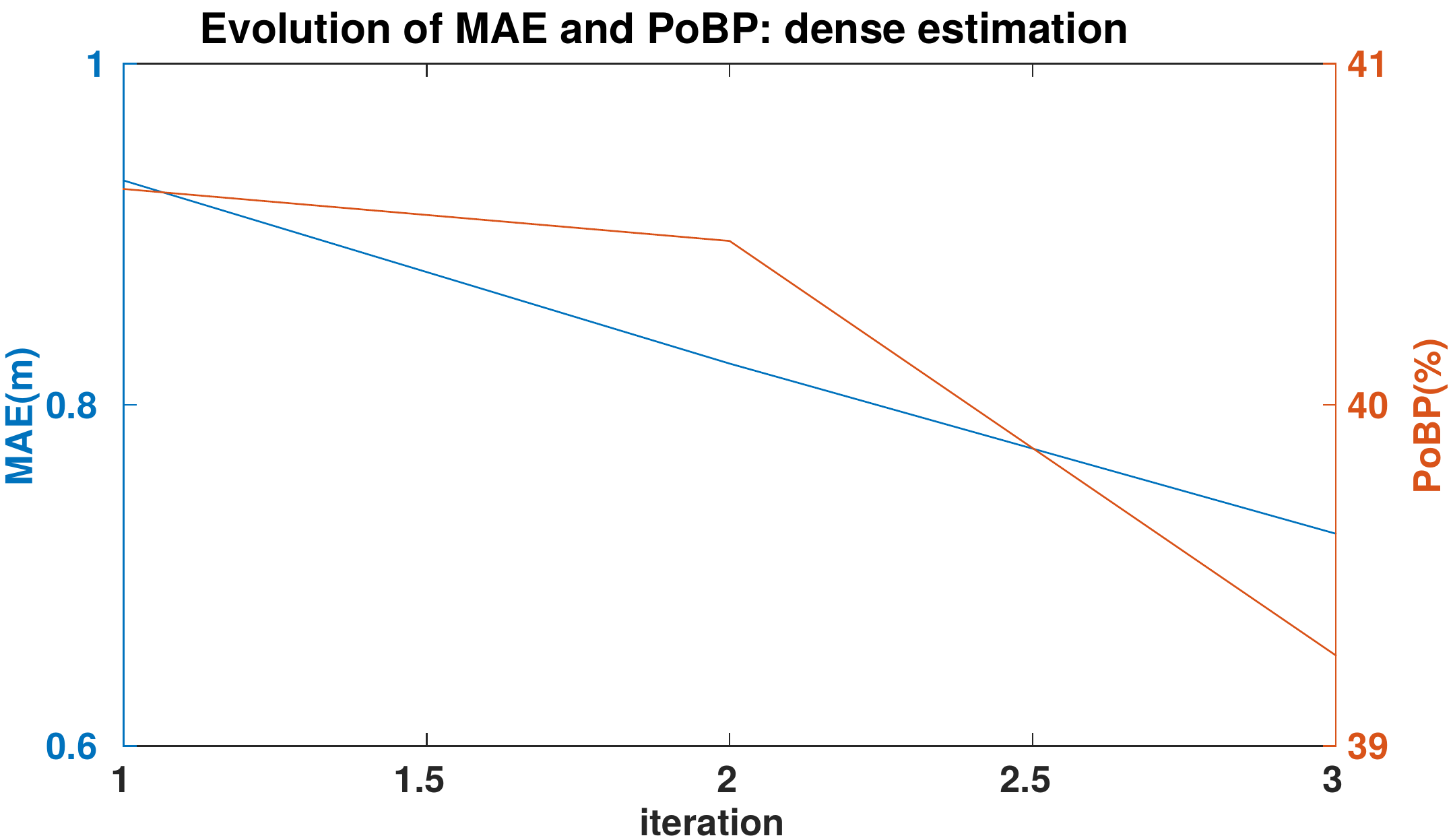}
	\end{minipage}
\end{center}
\vspace*{-4mm}
\caption{(Left) Sparse estimated 3D structure and ground-truth depth (dark blue represents closer objects and yellow the furthest ones); (right) evolution along iterative refinement of MAE (left in blue) and PoBP (right in red) for sparse and dense estimations. First row: frame 3 of Fountain sequence; remainder rows: (top) landscape image of estimated sparse depth and (bottom) sparse ground-truth for various example frames of the Kitti benchmark.}%
\label{fig:3Dstructure}
\end{figure*}

In this section we evaluate the accuracy of our method's 3D geometry estimation by comparing to the disparity ground-truth (subject to availability). To obtain the scale of the depth, we use the provided ground-truth translational motion speed. Two metrics are used: 1) Mean Absolute Error (MAE), and 2) Percentage of Bad Points (PoBP).

\begin{table}[]
\centering
\caption{3D structure error metrics: MAE and PoBP}
\label{tab:3Dstructure_error}

\scalebox{1.0}{
   \resizebox{1\textwidth}{1.cm}{
		\begin{tabular}{l||rr|rr|r}
		\hline
		 & MAE dense & PoBP dense & MAE sparse & PoBP sparse & density \\ \hline
		Yosemite sequence \cite{YOS_barron_2017} & NA & NA & NA & NA & 10.54\% \\
		Fountain sequence \cite{FOU_raudies_2017} & 0.359 & 15.60\% & 0.520 & 9.24\%~(1.44\%) & 9.60\% \\
		Kitti dataset \cite{geiger_ready_2012} & 0.800 & 32.91\% & 1.468 & 16.40\%~(1.80\%) & 10.95\% \\\hline
		\end{tabular}
	}
}
\end{table}
Table~\ref{tab:3Dstructure_error} shows the accuracy of the 3D structure using these two error metrics, in addition to the density values which indicate the sparseness of the estimation (common in sparse disparity estimation works such as \cite{brandt_improved_1997,barranco_parallel_2012}).
The first two columns provide the errors for the dense estimate after applying the inpainting method. The average error is below 0.4~m for the Fountain sequence (the maximum depth here is 7-8~m) and below 0.8~m for the Kitti dataset (with ranges up to 15-20~m). The third and fourth columns list the errors for our sparse estimate (about 10~\% of the image resolution) before applying the regularization. Comparing the results in columns 1-4, we see that, because the inpainting helps recovering smooth surfaces where no gradients are found, it reduces the average absolute error. On the other hand, it negatively affects the PoBP, since the reconstruction process propagates the error at object contours due to the absence of values nearby. The values in parenthesis provide the PoBP for the whole image, not only the sparse estimate, which for both datasets is below 2\%.

Fig.~\ref{fig:3Dstructure} illustrates the estimation of the 3D structure with examples from each sequence. The left and right image in row one, and top and bottom image in all other rows show the sparse estimated 3D structure and the sparse disparity ground-truth. The two plots on the right show the evolution of the MAE (in blue on the left axis) and the PoBP (in red on the right axis) along the refinement process. The number of iterations of this process depends on a convergence threshold that stops any further processing when there is not enough change between consecutive refined estimates. The first row shows frame 3 of the Fountain sequence, the second row shows frame 3 of the Yosemite sequence, and the last four rows show four frames from the Kitti dataset. As discussed in Table~\ref{tab:3Dstructure_error}, the regularization process for the reconstruction negatively affects the PoBP, while it reduces the average MAE over all pixels. In most cases each refinement step reduces the MAE, for example 50\% for sequence 5 of the Kitti dataset, while decreasing the PoBP.


Regarding the time performance, the bottleneck in our processing is the search for the translation. For the Fountain sequence, the average running time for the whole computation is 97.6~s, where 84.1~s correspond to the search, 13.05~s to the refinement (including three iterations until convergence), and the inpainting process requires 1.4~s. After parallelizing the search using 8 threads in a 4-GHz Intel Core i7 computer with 8 GB of RAM, the time for the search has been reduced to 13.04~s (and it is expected to be further reduced when using a massively parallel hardware such as a GPU). Comparing the time to optical flow based methods, we found the following: Using the same computer, Sun's method \cite{sun_quantitative_2014} requires 61.56~s, Brox's method \cite{brox_large_2011} requires 5.95~s, and Vogel's method \cite{vogel_evaluation_2013} 29.14~s. After that, the time for running the 3D motion estimation methods is 0.76~s for Raudies's \cite{raudies_review_2012} and 0.73~s for Kanatani's \cite{kanatani_3dinterpretation_1993}. However, as mentioned before, these methods rely on RANSAC for more refined estimations, and in this case Raudies's requires 97.81~s and Kanatani's about 200.81~s. Bearing in mind the accuracy of our direct method, the time performance is very similar to the conventional optical flow based methods.

\section{Conclusions}
\label{sec:6}
We have introduced a new formulation of the depth positivity constraint. On the basis of this constraint, we proposed a direct method that allows for joint estimation of 3D motion and 3D structure using as input image gradients. The complete method consists of a non-linear optimization for the positive depth constraint using normal flow, followed by a refinement using a linear optimization on the depth. We showed that our method obtains higher accuracy in motion estimation than other direct methods, and other optical flow based 3D motion estimation techniques. Furthermore, the estimated 3D geometry of the scene was shown of good quality.

Our results demonstrate that delaying the smoothness constraint, and estimating 3D motion globally in early stages of the structure-from-motion pipeline from minimal data, is a good choice. While we used the new depth positivity constraint with normal flow, it also could be applied to optical flow. Thus our work may inspire further approaches that include the positivity constraint into the 3D motion estimation. One possibility is to incorporate the constraint into a deep learning architecture for motion and structure estimation \cite{zhou2017unsupervised,yin2018geonet}.

\section*{Acknowledgment}
This work was supported by a \textit{Juan de la Cierva} grant (IJCI-2014-21376), partially funded by MINECO-FEDER TIN2016-81041-R grant, and the National Science Foundation under grants  SMA 1540917 and CNS 1544797.


\bibliography{mybibfile}

\end{document}